\documentclass[sn-basic]{sn-jnl}

\usepackage{graphicx}%
\usepackage{multirow}%
\usepackage{amsmath,amssymb,amsfonts}%
\usepackage{amsthm}%
\usepackage{mathrsfs}%
\usepackage[title]{appendix}%
\usepackage{xcolor}%
\usepackage{textcomp}%
\usepackage{manyfoot}%
\usepackage{booktabs}%
\usepackage{algorithm}%
\usepackage{algorithmicx}%
\usepackage{algpseudocode}%
\usepackage{listings}%

\usepackage{amsmath,amssymb,amsfonts}
\usepackage{algpseudocode,algorithm}
\usepackage{subcaption}
\usepackage[T1]{fontenc}
\usepackage{graphicx}
\usepackage{tabularx}
\usepackage{url}
\usepackage{orcidlink}

\theoremstyle{thmstyleone}%
\theoremstyle{thmstyletwo}%

\theoremstyle{thmstylethree}%

\begin{document}

\title[Revisiting Experience Replayable Conditions]{Revisiting Experience Replayable Conditions}


\author*[1]{\fnm{Taisuke} \sur{Kobayashi} \orcidlink{0000-0002-3760-249X}}\email{kobayashi@nii.ac.jp}

\affil*[1]{\orgname{National Institute of Informatics (NII) / The Graduate University for Advanced Studies (SOKENDAI)}, \orgaddress{\street{2-1-2 Hitotsubashi}, \city{Chiyoda-ku}, \postcode{101-8430}, \state{Tokyo}, \country{Japan}}}

\abstract{%
Experience replay (ER) used in (deep) reinforcement learning is considered to be applicable only to off-policy algorithms.
However, there have been some cases in which ER has been applied for on-policy algorithms, suggesting that off-policyness might be a sufficient condition for applying ER.
This paper reconsiders  more strict ``experience replayable conditions'' (ERC) and proposes the way of modifying the existing algorithms to satisfy ERC.
In light of this, it is postulated that the instability of policy improvements represents a pivotal factor in ERC.
The instability factors are revealed from the viewpoint of metric learning as i) repulsive forces from negative samples and ii) replays of inappropriate experiences.
Accordingly, the corresponding stabilization tricks are derived.
As a result, it is confirmed through numerical simulations that the proposed stabilization tricks make ER applicable to an advantage actor-critic, an on-policy algorithm.
Moreover, its learning performance is comparable to that of a soft actor-critic, a state-of-the-art off-policy algorithm.
}


\maketitle

\section{Introduction}

In the past decade, (deep) reinforcement learning (RL)~\citep{sutton2018reinforcement} has made significant progress and has gained attention in various application fields, including
game AI~\citep{oh2021creating,wang2021scc},
robot control~\citep{kalashnikov2018scalable,wu2023daydreamer},
autonomous driving~\citep{chen2021interpretable,cui2021autonomous},
and even finance~\citep{hambly2023recent}.
RL agents can optimize their action policies for unknown environments in a trial-and-error manner.
However, the trial-and-error process is computationally expensive, and the low sample efficiency of RL remains a challenge.

To improve sample efficiency, many RL algorithms utilize experience replay (ER)~\citep{lin1992self}.
ER stores the empirical data obtained during the trial-and-error process into a replay buffer instead of immediately streaming them.
By randomly and repeatedly replaying the stored empirical data for learning, the impact of each empirical data for learning, i.e. sample efficiency, can be increased.
Although ER is a simple technique, it is highly useful due to its compatibility with recent deep learning libraries and abundant computational resources.
However, it is claimed that ER is only applicable to off-policy algorithms~\citep{fedus2020revisiting}.
For example, proximal policy optimization (PPO), an on-policy algorithm, learns after collecting a certain amount of empirical data but generally does not reuse them~\citep{schulman2017proximal}.

Nevertheless, a survey of past cases suggests the efficacy of ER to on-policy algorithms.
For example, previous studies~\citep{zhao2016deep,bejjani2018planning} reported that ER was successfully applied to state–action–reward–state–action (SARSA)~\citep{sutton2018reinforcement}, a typical on-policy algorithm
\footnote{\citet{bejjani2018planning} modified SARSA to deep Q-networks (DQN)~\citep{mnih2015human}, an off-policy algorithm, in subsequent updates~\citep{bejjani2018planning2}.}.
This SARSA is also used to learn value functions in soft actor-critic (SAC)~\citep{haarnoja2018soft} and twin delayed deep deterministic policy gradient (TD3)~\citep{fujimoto2018addressing}, both of which are off-policy algorithms
\footnote{Theoretically, their learning rules corresponds to the expected SARSA~\citep{van2009theoretical}, which is an off-policy algorithm, but their implementations are with a rough Monte Carlo approximation and would lack rigor.}.
In addition, the author have successfully applied ER to PPO and its variants without any learning breakdowns~\citep{kobayashi2023proximal} (also see Appendix~\ref{app:ppo}).

From the above suggestive cases, it is expected that being an off-policy algorithm corresponds to a sufficient condition for ``experience replayable conditions'' (ERC).
Then, what is necessary and sufficient conditions for ERC?
This study hypothesizes that \textit{i) policy improvement algorithms in RL have the corresponding sets of empirical data that can be acceptable for learning}, and that \textit{ii) ER can be applied to them if only the empirical data belonging to the respective sets are replayed}.
With these hypotheses, one can expect that off-policy algorithms naturally satisfy ERC because the acceptable set for them coincides with the whole set of empirical data.
As a remark, based on the past cases, it is assumeed that there is no ERC in the context of learning value functions (this assumption will become clear in the numerical verification of this paper).

To demonstrate the validity of hypotheses, this paper first reveals the instability factors that destabilize learning by ER.
Specifically, one can focus on the fact revealed in~\citet{kobayashi2022optimistic} that the standard policy improvement algorithms can be derived as a triplet loss in metric learning~\citep{wang2014learning}, from the viewpoint of control as inference (CaI)~\citep{levine2018reinforcement}.
That is, the instability factors of triplet loss, i.e. i) repulsive forces from negative samples (so-called \textit{hard negative})~\citep{cheng2016person,xuan2020hard} and ii) replays of inappropriate experiences (so-called \textit{distribution shift})~\citep{schroff2015facenet,yu2018correcting}, are inherited by the policy improvement algorithms.
Since the way of sampling data makes the instability factors active, ER also inherit them by randomly selecting empirical data, causing learning breakdowns.

\begin{figure}[tb]
    \centering
    \includegraphics[keepaspectratio=true,width=0.96\linewidth]{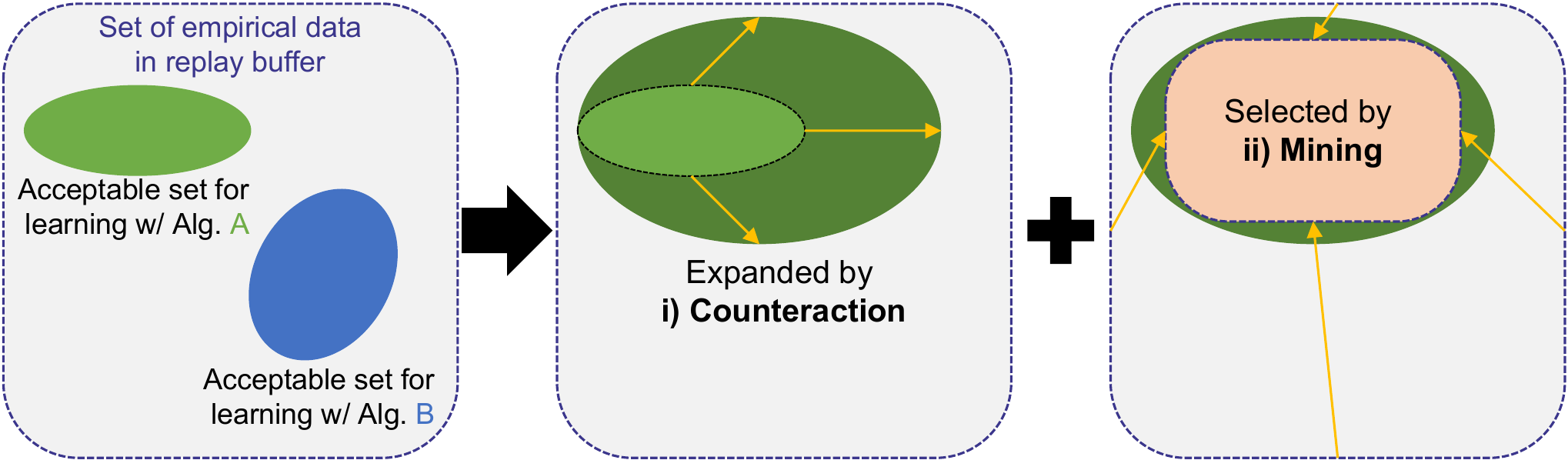}
    \caption{Concept of two stabilization tricks for satisfying ERC:
    Under the hypothesis that different algorithms have different sets of acceptable empirical data, the first trick, \textit{counteraction}, expands that set, while the second trick, \textit{mining}, selects the empirical data to be replayed.
    }
    \label{fig:concept_erc}
\end{figure}

To alleviate the identified instability factors, the two corresponding stabilization tricks, a \textit{counteraction} and a \textit{mining}, are proposed respectively based on an \textit{experience discriminator}.
For the hypothesized ERC, i) the counteraction is responsible for expanding the acceptable set of empirical data, and ii) the mining is responsible for selectively replaying the empirical data belonging to it (see Fig.~\ref{fig:concept_erc}).
Hence, ERC should hold by setting both of the counteraction and mining appropriately.

With the proposed stabilization tricks, it is verified whether ER is applicable to an advantage actor-critic (A2C)~\citep{mnih2016asynchronous}, an on-policy algorithm, by satisfying ERC.
First, a rough grid search of hyperparameters is performed to identify the range of counteraction and mining in which ERC holds.
Second, ablation tests for the counteraction and mining with the found hyperparameters show the needs for both.
Finally, the A2C with the proposed stabilization tricks is evaluated on more practical multi-dimensional tasks, suggesting the learning performance comparable to SAC~\citep{haarnoja2018soft}, which is the state-of-the-art off-policy algorithm.

In summary, this paper contributes to the following three folds:
\begin{enumerate}
    \item It is revealed that the reason why ER is limited to off-policy algorithms is due to the two instability factors associated with the triplet loss hidden in the policy improvement algorithms.
    \item The two stabilization tricks, the counteraction and mining, are developed to alleviate the corresponding instability factors respectively.
    \item With the two stabilization tricks, A2C, an on-policy algorithm, can satisfy ERC and achieve learning performance comparable to SAC.
\end{enumerate}

\section{Preliminaries}

\subsection{Basics of reinforcement learning}

Let's briefly introduce a problem statement of RL~\citep{sutton2018reinforcement}.
RL aims to maximize the sum of future rewards, so-called return, under Markov decision process (MDP).
Mathematically, MDP is defined with the tuple $(\mathcal{S}, \mathcal{A}, p_e, r)$:
$\mathcal{S}$ and $\mathcal{A}$ denote the state and action spaces
\footnote{Continuous spaces are assumed without loss of generality to match the experiments in this paper.}, respectively;
$p_e: \mathcal{S} \times \mathcal{A} \times \mathcal{S} \mapsto [0, \infty)$ gives the stochastic state transition function;
and $r: \mathcal{S} \times \mathcal{A} \mapsto \mathcal{R}$ defines the reward function for the current situation with $\mathcal{R} \subset \mathbb{R}$ the set of rewards.

Under the above MDP, an agent encounters the current state $s_t \in \mathcal{S}$ of an (unknown) environment at the current time step $t \in \mathbb{N}$.
The agent (stochastically) decides its action $a_t \in \mathcal{A}$ according to its learnable policy $\pi(a_t \mid s_t): \mathcal{S} \times \mathcal{A} \mapsto [0, \infty)$.
By interacting with the environment by $a_t$, the agent enters the next state $s_{t+1} =: s_t^\prime \in \mathcal{S}$ according to $p_e(s_t^\prime \mid s_t, a_t)$.
At the same time, the reward $r_t = r(s_t, a_t)$ is obtained from the environment.

By repeating the above transition, the agent obtains the return, $R_t = \sum_{k=0}^\infty \gamma^k r_{t+k}$, with $\gamma \in [0, 1)$ the discount factor.
RL wants to maximize the expected return from $t$ by optimizing $\pi(a_t \mid s_t)$ as follows:
\begin{align}
    \pi^\ast(a_t \mid s_t) = \arg\max_{\pi} \mathbb{E}_{a_t \sim \pi(a_t \mid s_t), \tau_t \sim \rho^\pi}[R_t \mid s_t]
    \label{eq:opt_policy}
\end{align}
where $\rho^\pi$ denotes the probability to generate the state-action trajectory, $\tau_t = [s_t^\prime, a_{t+1}, \ldots]$, with the combination of $\pi$ and $p_e$.
Here, the expected return can be defined as the following learnable value functions.
\begin{align}
    V(s_t) &= \mathbb{E}_{a_t \sim \pi(a_t \mid s_t), \tau_t \sim \rho^\pi}[R_t \mid s_t]
    \\
    Q(s_t, a_t) &= \mathbb{E}_{\tau_t \sim \rho^\pi}[R_t \mid s_t, a_t]
\end{align}
where $V(s_t) = \mathbb{E}_{a_t \sim \pi(a_t \mid s_t)}[Q(s_t, a_t)]$ holds.
The value functions are useful in acquring $\pi^\ast$, and therefore, most RL algorithms learn them.

\subsection{RL algorithms with experience replay}

ER stores the empirical data, $(s, a, s^\prime, r, b)$ ($b = \pi(a \mid s) \in \mathbb{R}_+$ the policy likelihood to generate $a$, which is used for the proposed method), in a replay buffer $\mathcal{B}$ with finite size $|\mathcal{B}|$~\citep{lin1992self}.
For the sake of simplicity, this paper focuses on ER for learning algorithms with each single transition, although ER for long-term sequences has also been developed~\citep{hansen2018fast,kapturowski2019recurrent}.
That is, the loss function w.r.t. the learnable functions in RL (i.e. $\pi$, $V$, and/or $Q$ mainly) to be minimized is required to be computed only with each empirical data.
The following minimization problem is therefore given under ER.
\begin{align}
    \min_{f} \mathbb{E}_{(s, a, s^\prime, r, b) \sim \mathcal{B}}[\ell_{f}^\mathrm{alg}(s, a, s^\prime, r, b)]
\end{align}
where $f$ indicates the function(s) to be optimized and $\ell_{f}^\mathrm{alg}$ denotes the algorithm-dependent loss function for $f$ (even if other functions are used for computing it, they are not optimized through minimizing it).
Although various variants have been proposed as the attention to practicality of ER has increased so far (e.g. prioritizing the important empirical data with high replay frequency~\citep{schaul2016prioritized,saglam2023actor} and ignoring the inappropriate empirical data~\citep{novati2019remember,sinha2022experience}), this paper employs the most standard implementation with the FIFO-type replay buffer and uniform sampling of empirical data.

\subsubsection{Soft actor-critic}

Here, two algorithms mainly used in this paper are introduced briefly.
The first one is SAC~\citep{haarnoja2018soft}, which is known as a representative off-policy algorithm, as the comparison.
SAC learns two functions, $\pi$ and $Q$, through minimization of the following loss functions
\footnote{The actual implementation has two $Q$ functions to conservatively compute the target value.}.
\begin{align}
    \ell_{Q}^\mathrm{SAC}(s, a, s^\prime, r) &= \frac{1}{2} \{ r + \gamma \mathbb{E}_{a^\prime \sim \pi(a^\prime \mid s^\prime)}[\bar{Q}(s^\prime, a^\prime) - \alpha \ln \pi(a^\prime \mid s^\prime)] - Q(s, a) \}^2
    \\
    \ell_{\pi}^\mathrm{SAC}(s) &= \mathbb{E}_{a \sim \pi(a \mid s)}[- Q(s, a) + \alpha \ln \pi(a \mid s)]
\end{align}
where $\bar{Q}$ denotes the target value (without computational graph) and $\alpha \geq 0$ denotes the magnitude of policy entropy, which can be auto-tuned by~\citet{kobayashi2023soft}.
The two expectations w.r.t. $\pi$ are roughly approximated by a one-sample Monte Carlo method.

As for the optimization of $Q$, the expected SARSA~\citep{van2009theoretical}, which can be an off-policy algorithm, is employed.
Indeed, the target value of $Q$ is computed with $\pi$, which is different from the policy used when obtaining the empirical data (so-called the behavior policy).
In addition, the optimization of $\pi$ is off-policy as well since $Q$ to be minimized in $\ell_{\pi}$ can be dependent on arbitrary policies (as like~\citet{lillicrap2016continuous}).
In this way, SAC belongs to the off-policy algorithm in total.
As a result, SAC can optimize $\pi$ and $Q$ with arbitrary empirical data, making it possible to use ER freely.

\subsubsection{Advantage actor-critic}
\label{subsubsec:a2c}

Next, A2C~\citep{mnih2016asynchronous}, an on-policy algorithm, is introduced as the baseline for the proposed method.
In this paper, the advantage function is approximated by the temporal difference (TD) error $\delta$ for simplicity.
In other words, A2C in this paper learns $\pi$ and $V$ by minimizing the following loss functions.
\begin{align}
    \delta(s, s^\prime, r) &= r + \gamma \bar{V}(s^\prime) - V(s)
    \nonumber \\
    \ell_{V}^\mathrm{A2C}(s, s^\prime, r) &= \frac{1}{2} \delta(s, s^\prime, r)^2
    \\
    \ell_{\pi}^\mathrm{A2C}(s, a, s^\prime, r) &= - \delta(s, s^\prime, r) \ln \pi(a \mid s)
\end{align}
where $\bar{V}$ denotes the target value as like $\bar{Q}$.

As $V$ necessarily depends on $\pi$ by definition, the above learning rule for $V$ requires $\pi$-dependent $r$ (and $s^\prime$).
The learning rule for $\pi$ is also derived based on the policy gradient theorem, and again the dependence of $\pi$ in $V$ is assumed.
In this way, A2C belongs to the on-policy algorithm in total.
As a result, A2C should not be theoretically able to learn with ER, which stores the empirical data independent of $\pi$.
Note that the policy gradient algorithm can theoretically be converted to the off-policy algorithm by utilizing importance sampling~\citep{degris2012off}, but due to computational instability and distribution shift, various countermeasures must be implemented~\citep{wang2017sample,zhang2019generalized,fakoor2020p3o} (see Section~\ref{subsec:onpolicy} in details).

\subsection{Metric learning with triplet loss}

\begin{figure}[tb]
    \centering
    \includegraphics[keepaspectratio=true,width=0.5\linewidth]{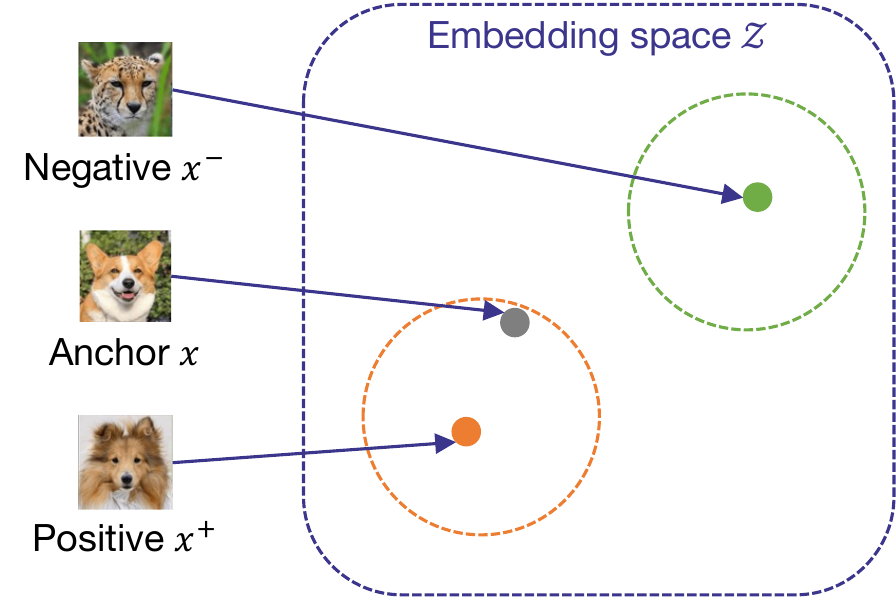}
    \caption{Desired distance relationship in metric learning with triplet loss:
    The anchor data $x$ should be in the same cluster as the positive data $x^+$, and the negative data $x^-$ should be taken from a different cluster away from it.
    }
    \label{fig:concept_triplet}
\end{figure}

Metric learning~\citep{bellet2022metric} is a methodology for extracting an embedding space, in which the similarity of the input data $x \in \mathcal{X}$ (e.g. image) can be measured.
The triplet loss relevant to this study~\citep{wang2014learning}, which is one of the loss functions to extract the embedding space $\mathcal{Z}$, is briefly introduced.
Three types of the input data, the anchor, positive, and negative data ($x$, $x^+$, and $x^-$ respectively), are required to compute it.
These are fed into the common networks $f$, outputting the corresponding features in $\mathcal{Z}$.
A distance function $d: \mathcal{Z} \times \mathcal{Z} \mapsto \mathbb{R}_+$ is then prepared to learn the desired distance relationship in the inputs.
That is, $x$ should be close to $x^+$ and away from $x^-$, as shown in Fig.~\ref{fig:concept_triplet}.

By minimizing the following triplet loss, this relationship can be acquired.
\begin{align}
    \ell_{f}^\mathrm{triplet} = \max(0, d(x, x^+) + m - d(x, x^-))
    \label{eq:def_triplet}
\end{align}
where $m \geq 0$ denotes the margin between the positive and negative clusters.
The max operator is employed to preclude the divergence into negative infinity.
As a result, the anchor data can be embedded near the positive data while being away from the negative data to some extent.

\section{Reasons for instabilization by experience replay}

\subsection{Policy improvements via control as inference}

CaI has been proposed for a new interpretation of RL~\citep{levine2018reinforcement}.
To indicate that the policy improvements (e.g. A2C) can be regarded as a kind of triplet loss, CaI is utilized as below (see~\citet{kobayashi2022optimistic} for the details of derivation).

In this concept, an optimal variable $O = \{0, 1\}$ is introduced.
The probability mass functions for it, $p(O=1 \mid s)$ and $p(O=1 \mid s, a)$, can be defined with $V$ and $Q$, respectively.
\begin{align}
    p(O=1 \mid s) &= e^{\beta (V(s) - C)} =: p_V(s)
    \label{eq:def_pv}\\
    p(O=1 \mid s, a) &= e^{\beta (Q(s, a) - C)} =: p_Q(s, a)
    \label{eq:def_pq}
\end{align}
where $\beta > 0$ denotes the inverse temperature parameter.
That is, if $\beta$ is small, these probabilities are likely to be near $1/2$, increasing ambiguity; and if $\beta$ is large, they often converge to $0$ or $1$, making them deterministic.
$C$ is an unknown value for convenience to satisfy $V(s) - C \leq 0$ and $Q(s, a) - C \leq 0$ so that the above equations satisfy the definition of probability and corresponds to the maximum value of the value function.
Since numerical computation is impossible with $C$ remaining, $C$ must be excluded from the learning rule that is eventually derived.
Note that since $O$ is binary, the probability of $O=0$ can be given as $p(O=0 \mid s) = 1 - p_V(s)$ and $p(O=0 \mid s, a) = 1 - p_Q(s, a)$.

Using these probabilities, the optimal and non-optimal policies are inferred using Bayesian theorem.
That is, $\pi(a \mid s, O)$ is given as follows:
\begin{align}
    \pi(a \mid s, O) &= \cfrac{p(O \mid s, a) b(a \mid s)}{p(O \mid s)}
    \nonumber \\
    &= \begin{cases}
        \cfrac{p_Q(s, a)}{p_V(s)} b(a \mid s) =: \pi^+(a \mid s) & O = 1
        \\
        \cfrac{1 - p_Q(s, a)}{1 - p_V(s)} b(a \mid s) =: \pi^-(a \mid s)& O = 0
    \end{cases}
\end{align}
where $b(a \mid s)$ denotes the behavior policy to sample $a$, which can be different from the current learnable policy $\pi$.

Now, to optimize $\pi$, the following minimization problem is solved.
\begin{align}
    \min_\pi \mathbb{E}_{s \sim \mathcal{S}}[\mathrm{KL}(\pi(a \mid s) \mid \pi^+(a \mid s)) - \mathrm{KL}(\pi(a \mid s) \mid \pi^-(a \mid s))]
    \label{eq:cai_triplet}
\end{align}
where $\mathrm{KL}(\cdot \mid \cdot)$ denotes Kullback-Leibler (KL) divergence between two probability distributions.
The gradient of these two terms w.r.t. $\pi$, $g_\pi$, is derived as follows:
\begin{align}
    g_\pi &=
    \mathbb{E}_{s \sim \mathcal{S}} \left[ \nabla \mathbb{E}_{a \sim \pi(a \mid s)} \left[ \ln \cfrac{\pi(a \mid s)}{\pi^+(a \mid s)} - \ln \cfrac{\pi(a \mid s)}{\pi^-(a \mid s)} \right] \right]
    \nonumber \\
    &= \mathbb{E}_{s \sim \mathcal{S}, a \sim \pi(a \mid s)}
    \left[ - \nabla \ln \pi(a \mid s) \left\{ \beta(Q(s, a) - V(s)) + \ln \cfrac{1 - p_Q(s, a)}{1 - p_V(s)} \right\} \right]
    \nonumber \\
    &\propto \mathbb{E}_{(s, a) \sim \mathcal{B}} \left [ - (Q(s, a) - V(s)) \nabla \ln \pi(a \mid s) \right ]
    \nonumber \\
    &\propto \mathbb{E}_{(s, a, s^\prime, r) \sim \mathcal{B}} \left [ - \delta(s, s^\prime, r) \nabla \ln \pi(a \mid s) \right ]
    \label{eq:cai_a2c}
\end{align}
Here, by assuming $\beta \to \infty$, this gradient can be simplified.
In addition, $Q - V = A$ the advantage function can be approximated by TD error $\delta$ (see Section~\ref{subsubsec:a2c}).
As mentioned above, when $\beta$ is large, $p(O=1)$ deterministically takes $0$ or $1$, and with $\beta \to \infty$, $O=1$ is not obtained unless $V = Q = C$ (i.e. the value function is maximized by the optimized policy, which is consistent with eq.~\eqref{eq:opt_policy}).

The minimization of eq.~\eqref{eq:cai_triplet} using the surrogated gradient is consistent with the minimization of the loss function for A2C, $\ell_\pi^\mathrm{A2C}$, by (stochastic) gradient descent.
The original minimization problem can be interpreted as trying to move the anchor data $\pi$ closer to the positive data $\pi^+$ and away from the negative data $\pi^-$ by employing KL divergence as the distance function $d$
\footnote{Although KL divergence does not actually satisfy the definition of distance, it is widely used in probability geometry because of its distance-like property, which is non-negative and zero only when two probability distributions coincide.}.
Compared to eq.~\eqref{eq:def_triplet}, the margin $m$ and the max operator are not used, but this is because $m=0$ and $\pi^{+,-}$ is centered on $b$, preventing the divergence to infinity.

\subsection{Instability factors hidden in triplet loss}

As the policy improvements correspond to minimizing triplet loss, then its characteristics during learning should also inherit that of minimizing triplet loss.
Ideally, the anchor data should approach the positive data, resulting in $\pi \to \pi^+$.
On the other hand, the minimization of triplet loss is different from simple supervised learning problems, and several factors that destabilize learning have been reported.
These instability factors are influenced by the way triplets are selected from the dataset.
In the policy improvements, therefore, they are activated by randomly sampling empirical data from ER (and using them for learning).
Under this assumption, the following two instability factors are raised, along with their solution guidelines noted in the context of metric learning, which should be useful for RL combined with ER.

First, selecting the inappropriate anchor, positive, and negative data might cause $\mathrm{KL}(\pi \mid \pi^+) > \mathrm{KL}(\pi \mid \pi^-)$.
In this case, the repulsion from $\pi^-$ is stronger than the attraction to $\pi^+$ and the optimal solution cannot be found, updating $\pi$ by the divergent behavior.
This is known as \textit{hard negative}~\citep{cheng2016person,xuan2020hard}.
To alleviate this instability factor, the exclusion of hard-negative triplets and/or the regularization that suppresses the repulsion would be required.

Second, from all the triplets that can be constructed, only a few can be used for optimization.
In fact, $\pi^+$ and $\pi^-$ are linked via the behavior policy $b$ in the policy improvements, and no arbitrary triplet can be constructed.
This might cause \textit{distribution shift}, inducing a large bias in learning~\citep{schroff2015facenet,yu2018correcting}.
To alleviate this instability factor, it is desirable to eliminate triplets that are prone to bias and/or to regularize the distribution of selected triplets.

\section{Tricks for experience replayable conditions}

\subsection{Experience discriminator}

\begin{figure}[tb]
    \begin{subfigure}[b]{0.48\linewidth}
        \centering
        \includegraphics[keepaspectratio=true,width=\linewidth]{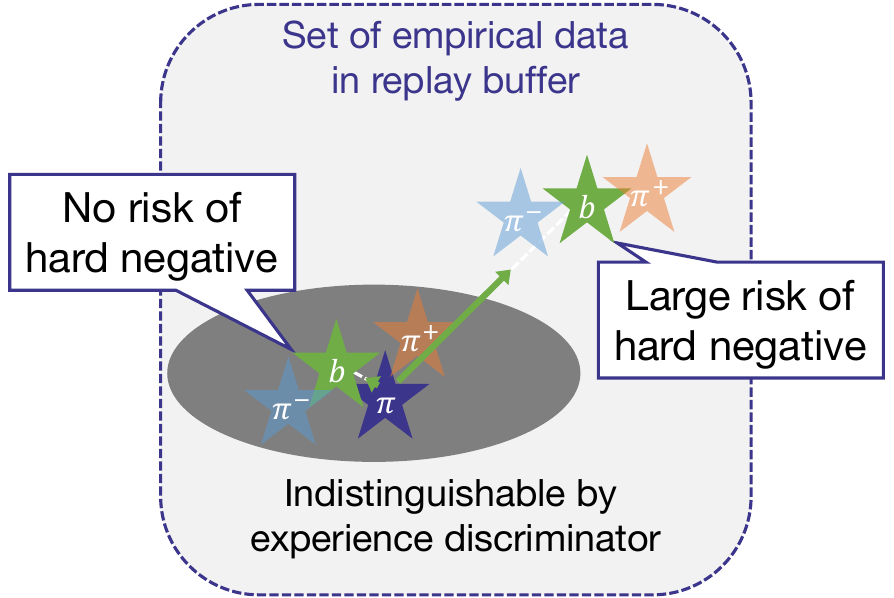}
        \subcaption{Counteraction}
        \label{fig:concept_counteraction}
    \end{subfigure}
    \begin{subfigure}[b]{0.48\linewidth}
        \centering
        \includegraphics[keepaspectratio=true,width=\linewidth]{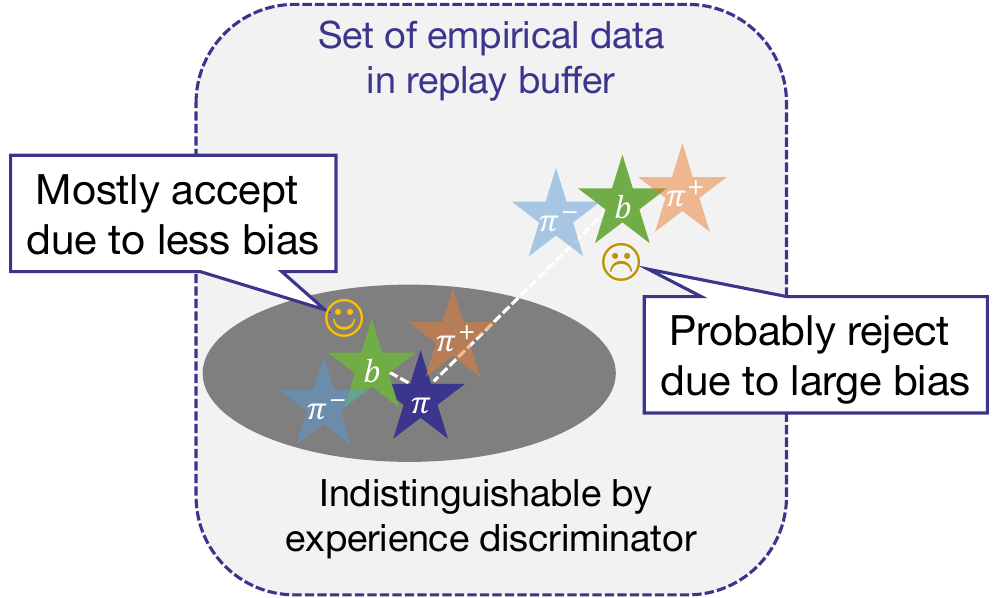}
        \subcaption{Mining}
        \label{fig:concept_mining}
    \end{subfigure}
    \caption{Two stabilization tricks:
    According to the judgements of \textit{experience discriminator},
    (a) \textit{counteraction} applies a regularization to $\pi$ to increase the misidentification rate to $\pi \simeq b$;
    and (b) \textit{mining} masks empirical data judged to be $\pi \neq b$ and excludes them from replay/training.
    }
    \label{fig:stabilization_tricks}
\end{figure}

The instability factors induced by the above triplet loss have to be suppressed to satisfy ERC.
To this end, two stabilization tricks, \textit{counteraction} and \textit{mining}, are heuristically designed (see Fig.~\ref{fig:stabilization_tricks}).
As a common module for them, \textit{experience discriminator}, is first introduced.
In other words, it estimates whether the empirical data of ER is suitable for triplet construction.

Specifically, it is required to judge whether the state-action pair $(s, a)$ in the buffer $\mathcal{B}$ can be regarded as one generated by the current policy $\pi$.
According to the density trick, the following density ratio $d$ satisfies this requirement.
\begin{align}
    d(s, a, b) &= \min \left( 0.5, \cfrac{\pi(a \mid s)}{\pi(a \mid s) + b} \right)
    \nonumber \\
    &= \min(0.5, \sigma(\ln \pi(a \mid s) - \ln b))
    \label{eq:density_ratio}
\end{align}
where $\sigma(\cdot) \in [0, 1]$ is the sigmoid function.
Note that $d$ should be equal to or less than $0.5$ since $(s, a)$ is actually generated from the behavior policy with its likelihood $b$.

In addition, robust judgements should be considered w.r.t. the stochasticity of actions and the non-stationarity of policies.
For this purpose, $D:\mathcal{S} \mapsto [0, 1]$, which marginalizes $d$ by $a$ and has only $s$ as input, is defined as a learnable model.
As $D$ corresponds to the probability parameter of Bernoulli distribution, it can be optimized by minimizing its negative log-likelihood.
\begin{align}
    \mathcal{L}_{D} = \mathbb{E}_{(s, a, b) \sim \mathcal{B}}[- d(s, a, b) \ln D(s) - (1 - d(s, a, b)) \ln (1 - D(s))]
    \label{eq:loss_discriminator}
\end{align}
where the above $d$ is employed as the supervised signal.

\subsection{Counteraction of deviations from non-optimal policies}

First trick, so-called the counteraction, is proposed as a countermeasure against the hard negative mainly.
In the previous work~\citep{cheng2016person}, the regularization to change the ratio of two terms in eq.~\eqref{eq:def_triplet} has been proposed to counteract the repulsion from the negative data.
This concept can be reproduced in eq.~\eqref{eq:cai_triplet} as follows:
\begin{align}
    \min_\pi \mathbb{E}_{s \sim \mathcal{S}}[(1 + \lambda)\mathrm{KL}(\pi(a \mid s) \mid \pi^+(a \mid s)) - \mathrm{KL}(\pi(a \mid s) \mid \pi^-(a \mid s))]
\end{align}
where $\lambda \geq 0$ denotes the gain for imbalancing the positive and negative terms
\footnote{The same is true if the second term is multiplied by the gain $\lambda \in (0, 1)$.}.

Now, how this extension works in theory is shown.
As in the original minimization problem, the gradient for the added regularization is derived as below.
\begin{align}
    & \nabla \mathbb{E}_{s \sim \mathcal{S}}[\mathrm{KL}(\pi(a \mid s) \mid \pi^+(a \mid s))]
    \nonumber \\
    =& \nabla \mathbb{E}_{s \sim \mathcal{S}, a \sim \pi}[\ln \pi(a \mid s) - \ln b(a \mid s) - \ln p_Q(s, a) + \ln p_V(s)]
    \nonumber \\
    =& \nabla \mathbb{E}_{s \sim \mathcal{S}}[\mathrm{KL}(\pi(a \mid s) \mid b(a \mid s))]
    - \nabla \mathbb{E}_{s \sim \mathcal{S}, a \sim \pi}[\ln p_Q(s, a) - \ln p_V(s)]
\end{align}
Now, the second term can be decomposed to eq.~\eqref{eq:cai_a2c} by using the definitions of $p_V$ and $p_Q$ in eqs.~\eqref{eq:def_pv} and~\eqref{eq:def_pq}, respectively.
\begin{align}
    & - \nabla \mathbb{E}_{s \sim \mathcal{S}, a \sim \pi}[\ln p_Q(s, a) - \ln p_V(s)]
    \nonumber \\
    =& - \nabla \mathbb{E}_{s \sim \mathcal{S}, a \sim \pi}[\beta Q(s, a) - \beta C - \beta V(s) + \beta C]
    \nonumber \\
    =& \mathbb{E}_{s \sim \mathcal{S}, a \sim \pi}[- \nabla \ln \pi(a \mid s) \beta \{Q(s, a) - V(s)\}]
    \nonumber \\
    \propto & \mathbb{E}_{(s, a, s^\prime, r) \sim \mathcal{B}} \left [ - \delta(s, s^\prime, r) \nabla \ln \pi(a \mid s) \right ]
    = \eqref{eq:cai_a2c}
\end{align}
That is, the second term can be absorbed into the original loss.

The added regularization, therefore, has a role to constrain $\pi \to b$ only.
As $b$ should be located between $\pi^+$ and $\pi^-$, this regularization is expected to avoid the hard negative and stabilize $\pi \to \pi^+$.
However, to implement this regularization directly, the behavior policy $b(a \mid s)$ at each experience must be retained, which is costly.
As an alternative regularization way, the adversarial learning to eq.~\eqref{eq:loss_discriminator} is implemented in this paper.

That is, it takes advantage of the fact that the higher the misidentification rate of $D$ means $\pi \sim b$, reducing the risk of hard negative (see Fig.~\ref{fig:concept_counteraction}).
Since $d$ has the computational graph w.r.t. $\pi$, the following loss function can be given as the counteraction trick for regularizing $\pi$ to $b$.
\begin{align}
    \mathcal{L}_{C} = \mathbb{E}_{(s, a, b) \sim \mathcal{B}}[\omega d(s, a, b) \{ \ln D(s) - \ln (1 - D(s)) \}]
    \label{eq:loss_counteraction}
\end{align}
where $\omega \geq 0$ denotes the gain, which is designed below.
One can find that the term related to $D$ also behaves as a gain and is larger when $D \ll 1$ (i.e. no misidentification).
In addition, if $d$ is clipped to $0.5$ (i.e. $\pi \simeq b$ holds), the gradient w.r.t. $\pi$ is zero.
Note that, in the actual implementation, the gradient reversal layer~\citep{ganin2016domain} is useful to lump eqs.~\eqref{eq:loss_discriminator} and~\eqref{eq:loss_counteraction} together.

As for $\omega$, $d \ll 1$ has the small gradient due to the sigmoid function, not like $D \ll 1$.
To compensate it and converge to $\pi \to b$ faster even in such a case, $\omega$ is designed in the manner of PI control, referring to the literature~\citep{stooke2020responsive}.
\begin{align}
    \begin{split}
        P(s, a, b) &= \eta_C (1 - 2 \hat{d}(s, a, b))
        \\
        I &= \max(0, 0.5I + \mathbb{E}_{\mathcal{B}}[P(s, a, b)])
        \\
        \omega(s, a, b) &= \max(0, P(s, a, b) + I)
    \end{split}
\end{align}
where $\hat{d}$ means $d$ without the computational graph, and $\eta_C \geq 0$ denotes the hyperparameter for this counteraction.
$I \geq 0$ is the integral term with saturation by multiplying $0.5$ (its initial value must be zero).

\subsection{Mining of indistinguishable experiences}

Second trick, so-called the mining, is proposed to mitigate the effect of distribution shift mainly.
In the previous studies, semi-hard triplets, in which $d(x, x^+) + m < d(x, x^-)$ holds, are considered useful~\citep{schroff2015facenet}.
Since the optimization problem in this study sets $m=0$, $\mathrm{KL}(\pi \mid \pi^+) < \mathrm{KL}(\pi \mid \pi^-)$ seems better.
However, this is also regarded to be hard triplets, which would induce the selection bias.
Of course, $\mathrm{KL}(\pi \mid \pi^+) > \mathrm{KL}(\pi \mid \pi^-)$ is still not suitable for learning because of the hard negative relationship described above.
Hence, the triplets with $\mathrm{KL}(\pi \mid \pi^+) \simeq \mathrm{KL}(\pi \mid \pi^-)$ are desired to be mined.

Specifically, it takes advantage of the fact that, in this study, the anchor data is determined by the current policy $\pi$, and the positive and negative data, $\pi^+$ and $\pi^-$, are located around the behavior policy $b$.
In other words, the indistinguishable empirical data with $\pi \simeq b$ is likely to achieve the desired relationship (see Fig.~\ref{fig:concept_mining}).
The mining trick is therefore given as the following stochastic dropout~\citep{srivastava2014dropout}.
\begin{align}
    \begin{split}
        p(M \mid s, a, b) &= 2\max(0, 0.5 - \min(d(s, a, b), D(s))^{\eta_M})
        \\
        M &= \begin{cases}
            1 & p(M \mid s, a, b) \leq \epsilon \sim \mathcal{U}(0, 1)
            \\
            0 & \text{otherwise}
        \end{cases}
    \end{split}
\end{align}
where $\mathcal{U}(l, u)$ is the uniform distribution within $[l, u]$, and $\eta_M \geq 0$ denotes the hyperparameter for this mining.
When sampling the empirical data from the replay buffer $\mathcal{B}$, each data is screend by the mining: if $M=1$, it is used for the optimization; if $M=0$, it is excluded.

\section{Simulations}

\subsection{Overview}

Here, it is verified that ERC holds by the two stabilization tricks.
To do so, A2C~\citep{mnih2016asynchronous}, which is the on-policy algorithm and generally considered inapplicable for ER, is employed as a baseline (see Appendix~\ref{app:impl} for detailed settings).
Learning is conducted only with the empirical data replayed from ER at the end of each episode (without online learning).

The following three steps are performed for the step-by-step verification.
With them, it is shown that alleviating the instability factors hidden in triplet loss is effective not only to satisfy ERC, but also to learn the optimal policy with comparable sample efficiency to SAC~\citep{haarnoja2018soft}.
\begin{enumerate}
    \item The hyperparameters of the respective stabilization tricks, $\eta_{C,M} \geq 0$, are roughly examined their effective ranges in a toyproblem, determining the values to be used in the subsequent verification.
    \item It is confirmed that both of the two stabilization tricks are complementarily essential to satisfy ERC and to learn the optimal policy through ablation tests on three major tasks implemented in Mujoco~\citep{todorov2012mujoco}.
    \item Finally, the A2C with the proposed stabilization tricks is evaluated on two tasks of relatively large problem scale in dm\_control~\citep{tunyasuvunakool2020dm_control} in comparison to SAC, which is the latest off-policy algorithm.
\end{enumerate}
Note that training for each task and condition is conducted 12 times with different random seeds in order to evaluate the performance with their statistics.
In the above, the stabilization tricks demonstrate that ER can be applied to A2C with the excellent learning performance.
In addition, the application of the two stabilization tricks to the other on-policy algorithm, i.e. PPO~\citep{schulman2017proximal}, and their quantitative contribution to the learning performance are investigated in Appendix~\ref{app:ppo}.

\subsection{Effective range of hyperparameters}

\begin{figure}[tb]
    \centering
    \includegraphics[keepaspectratio=true,width=0.8\linewidth]{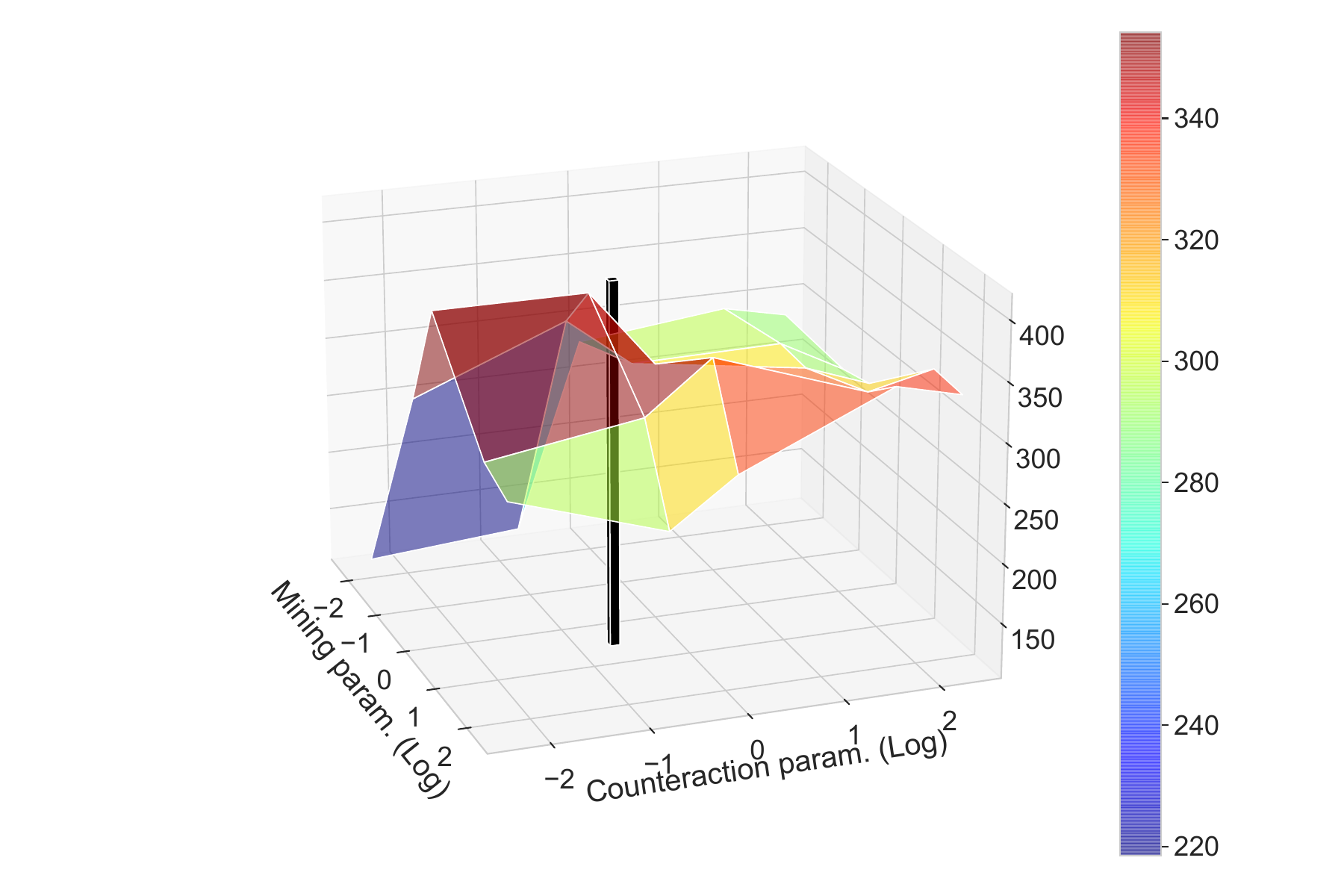}
    \caption{Grid search of $(\eta_C, \eta_M)$ on \textit{DoublePendulum}:
    $\eta_{C,M} = \{0.1, 0.5, 1.0, 5.0, 10.0\}$ are searched roughly in terms of return from the environment,
    and then $(\eta_C, \eta_M)$ used later are decided to be $(0.5, 2.0)$ referring to the grid search results.
    }
    \label{fig:erc_param_3d}
\end{figure}

A toyproblem named \textit{DoublePendulum}, where a pendulum agent with two passive joints tries to balance in a standing position by moving its base, is solved.
This has only one-dimensional action space and the state space limited by a terminal condition (i.e. excessive tilting).
Thus, as this is a relatively simple problem and many empirical data should naturally satisfy ERC, the effect of $\eta_{C,M}$ should behave in a Gaussian manner, enabling adjustment statistically.

To roughly check the effective parameter ranges of $\eta_{C,M}$, $5 \times 5 = 25$ conditions with $\eta_{C,M} = \{0.1, 0.5, 1.0, 5.0, 10.0\}$ are compared.
The test results, which are evaluated after learning, are summarized in Fig.~\ref{fig:erc_param_3d}.
The results suggest the following two points.

First, $\eta_M$ must be of a certain scale to activate the selection of empirical data, or else the performance can be significantly degraded.
However, an excessively large $\eta_M$ would result in too little empirical data being replayed.
Therefore, $\eta_M$ is considered reasonable in the vicinity of $1$.

Second, $\eta_C$ seems to have an appropriate range depending on $\eta_M$.
In other words, if $\eta_M$ is excessively large and the replayable empirical data is too limited, it is desirable to increase them by increasing $\eta_C$.
On the other hand, if the empirical data are moderately selected around $\eta_M \simeq 1$, it seems essential to relax the counteraction to some extent as $\eta_C \leq 1$.

Based on the above two trends, and as a natural setting, $(\eta_C, \eta_M)$ are decided to be $(0.5, 2.0)$ for the remaining experiments.
The learning performance at this time is also shown in Fig.~\ref{fig:erc_param_3d} (the black bar), and is higher than the rough grid search results (i.e. $\sim 416$ while the others did not exceed $400$).

\subsection{Ablation tests}

\begin{figure}[tb]
    \begin{subfigure}[b]{0.32\linewidth}
        \centering
        \includegraphics[keepaspectratio=true,width=\linewidth]{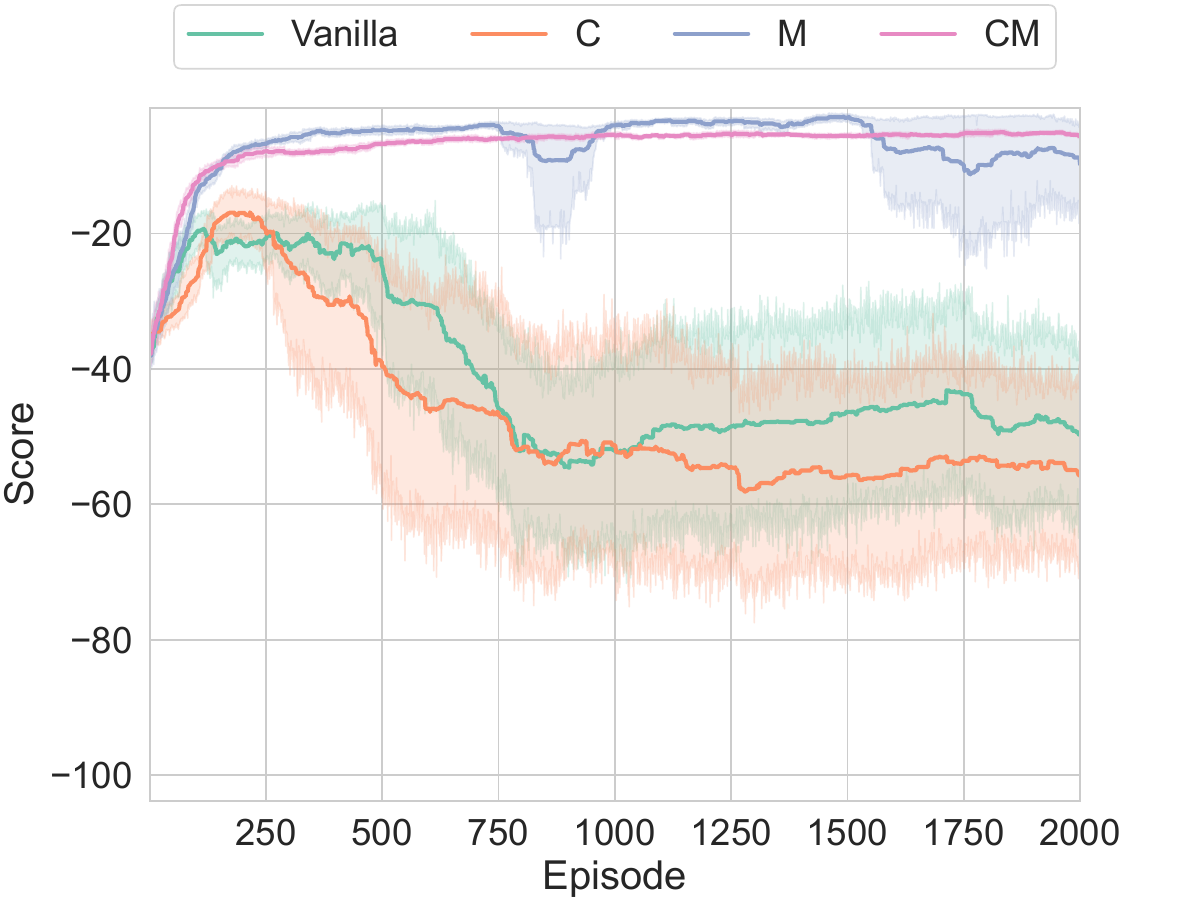}
        \subcaption{Reacher}
        \label{fig:score_Reacher}
    \end{subfigure}
    \begin{subfigure}[b]{0.32\linewidth}
        \centering
        \includegraphics[keepaspectratio=true,width=\linewidth]{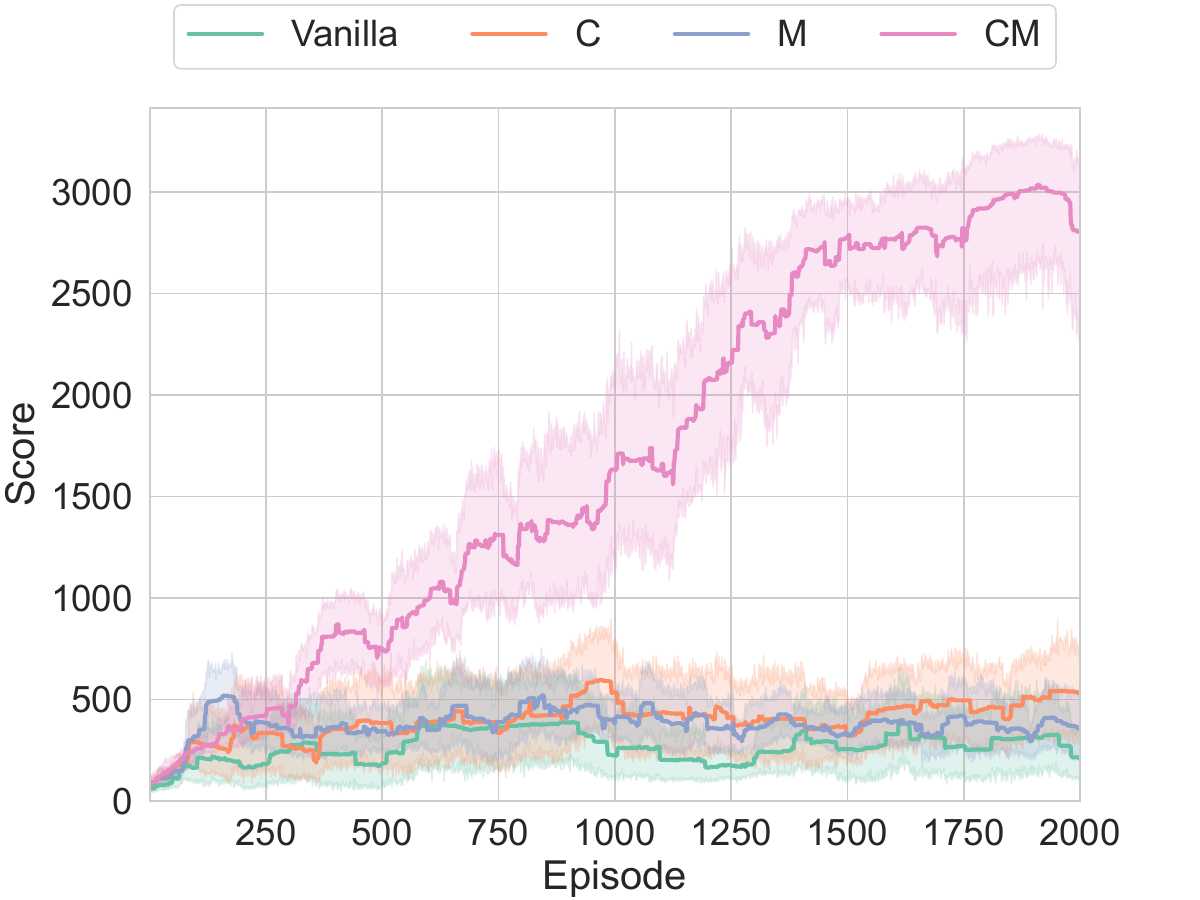}
        \subcaption{Hopper}
        \label{fig:score_Hopper}
    \end{subfigure}
    \begin{subfigure}[b]{0.32\linewidth}
        \centering
        \includegraphics[keepaspectratio=true,width=\linewidth]{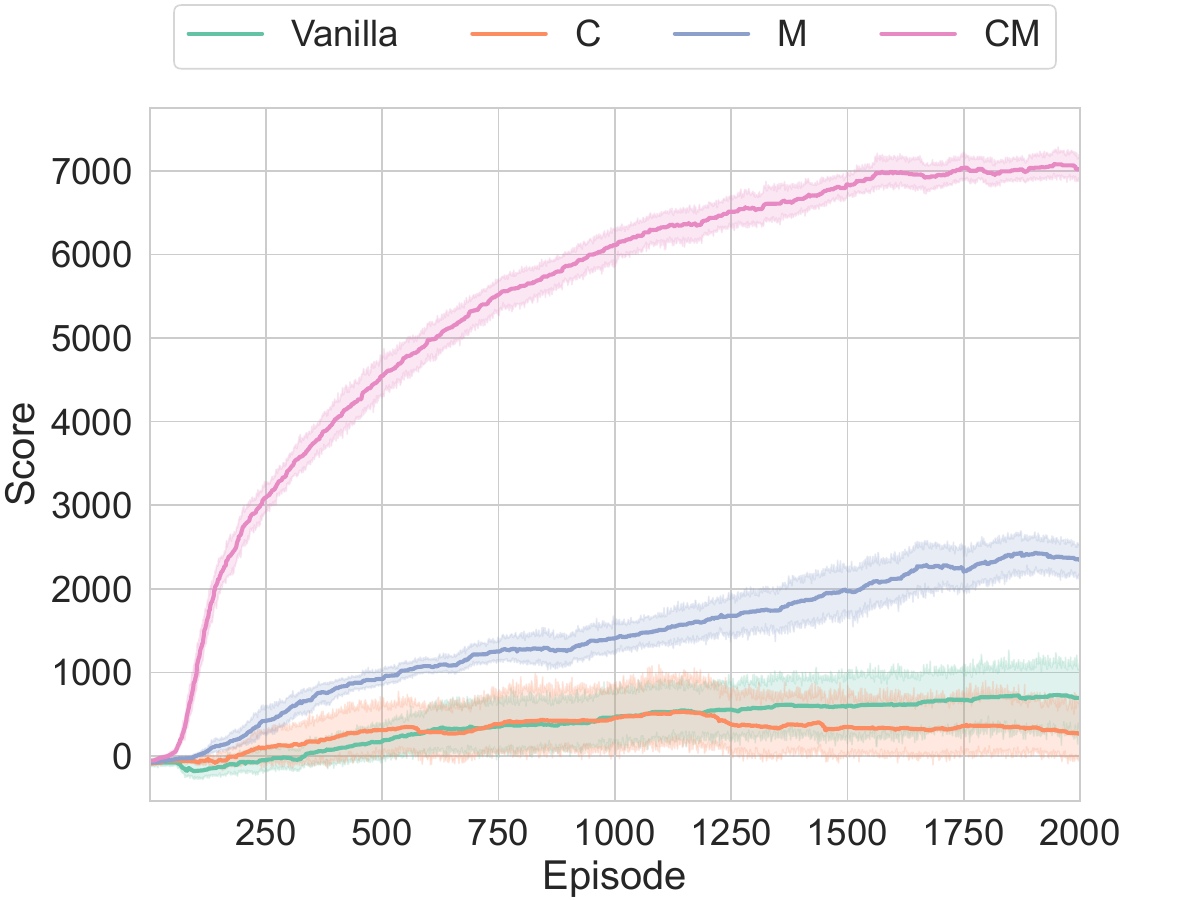}
        \subcaption{HalfCheetah}
        \label{fig:score_HalfCheetah}
    \end{subfigure}
    \caption{Returns of ablation tests:
    The addition of the two stabilization tricks enabled the learning with an on-policy algorithm, A2C, combined with ER.
    }
    \label{fig:score}
\end{figure}
\begin{figure}[tb]
    \begin{subfigure}[b]{0.32\linewidth}
        \centering
        \includegraphics[keepaspectratio=true,width=\linewidth]{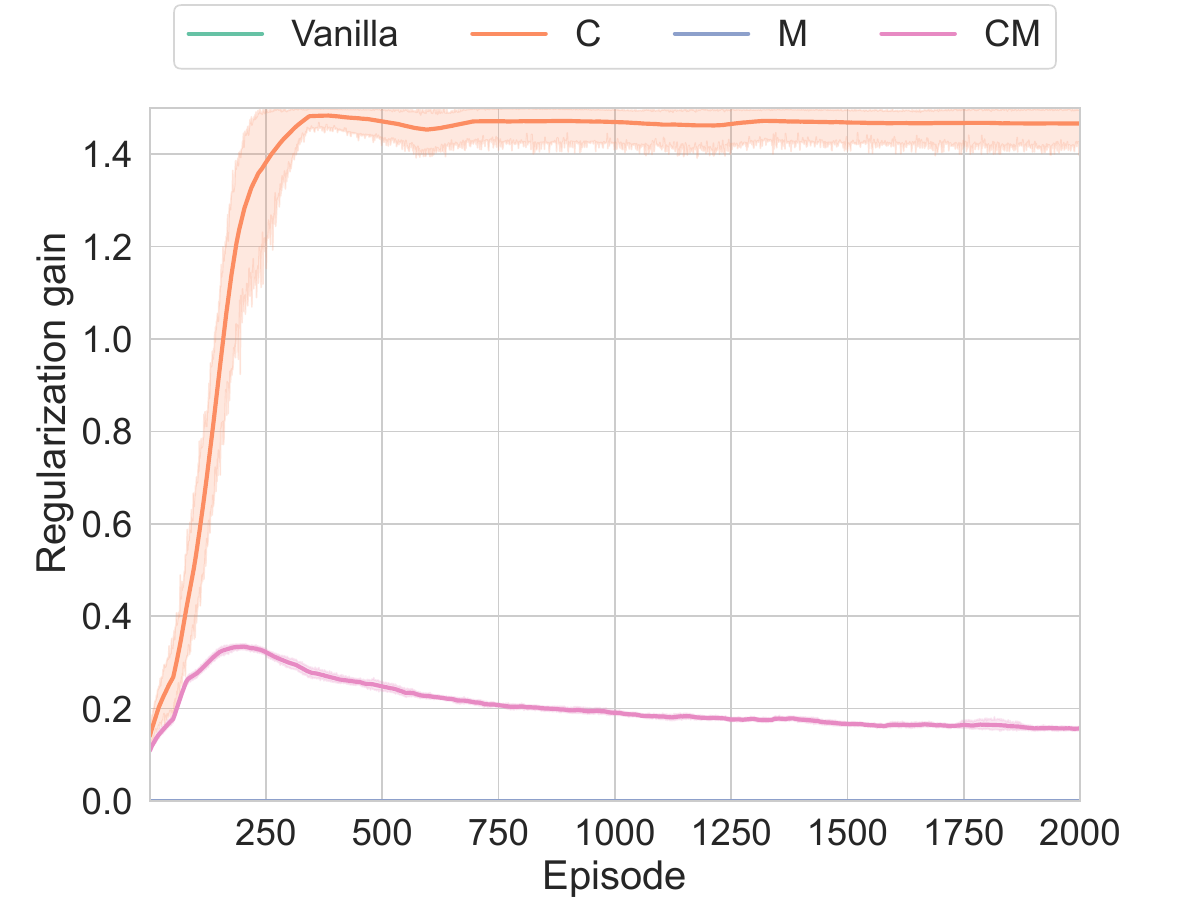}
        \subcaption{Reacher}
        \label{fig:erc_counteract_Reacher}
    \end{subfigure}
    \begin{subfigure}[b]{0.32\linewidth}
        \centering
        \includegraphics[keepaspectratio=true,width=\linewidth]{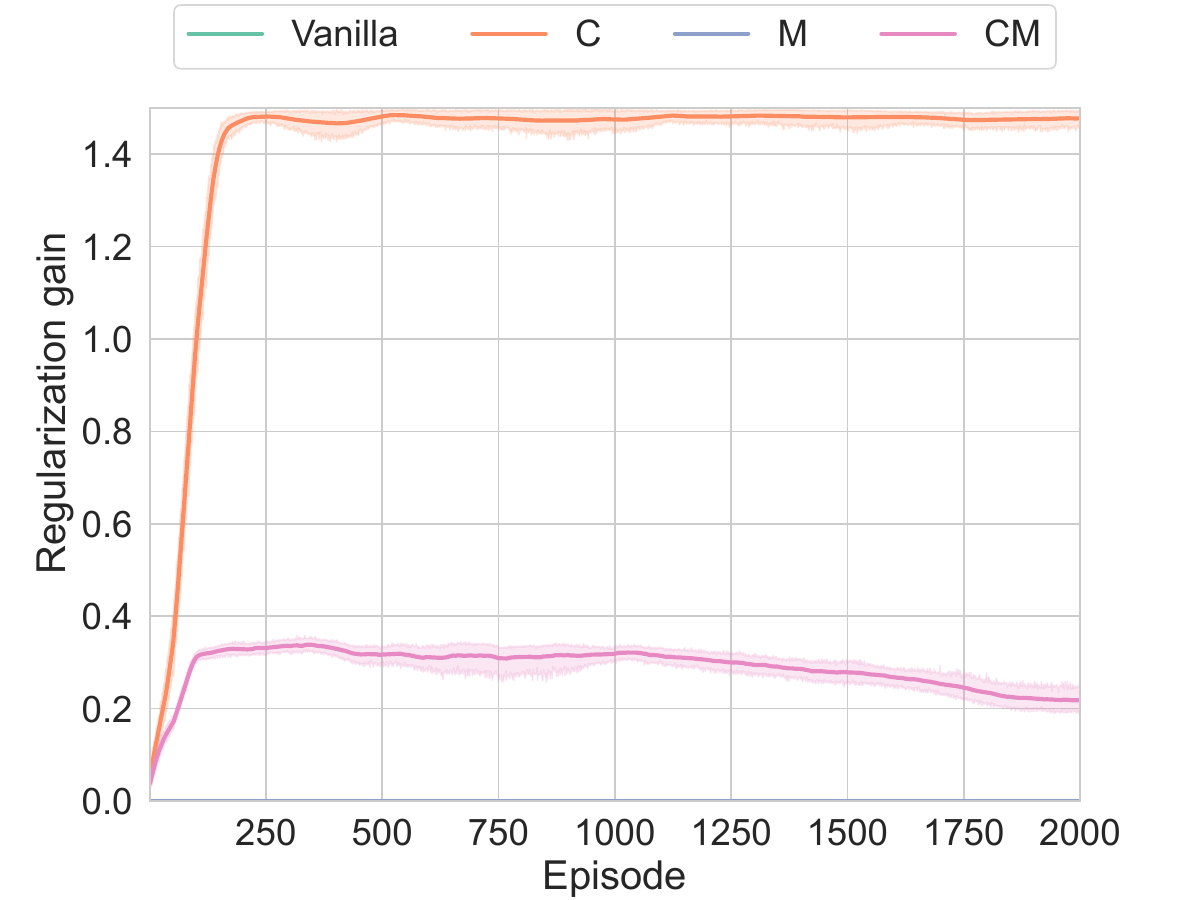}
        \subcaption{Hopper}
        \label{fig:erc_counteract_Hopper}
    \end{subfigure}
    \begin{subfigure}[b]{0.32\linewidth}
        \centering
        \includegraphics[keepaspectratio=true,width=\linewidth]{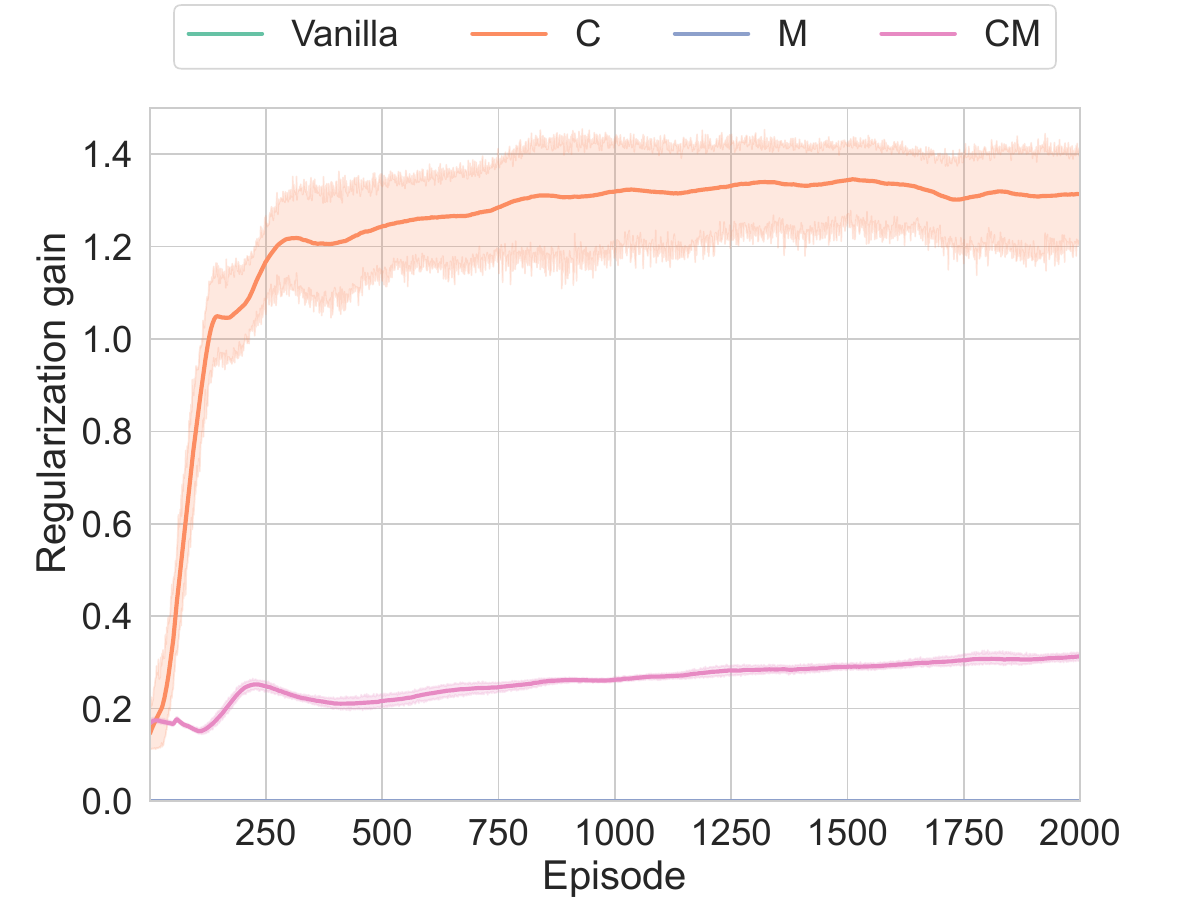}
        \subcaption{HalfCheetah}
        \label{fig:erc_counteract_HalfCheetah}
    \end{subfigure}
    \caption{$\omega(s, a, b)$ in the counteraction trick:
    While the counteraction trick alone saturated the regularization gain and made it difficult to perform $\pi \simeq b$, the two tricks made the regularization work properly without saturation.
    }
    \label{fig:erc_counteract}
\end{figure}
\begin{figure}[tb]
    \begin{subfigure}[b]{0.32\linewidth}
        \centering
        \includegraphics[keepaspectratio=true,width=\linewidth]{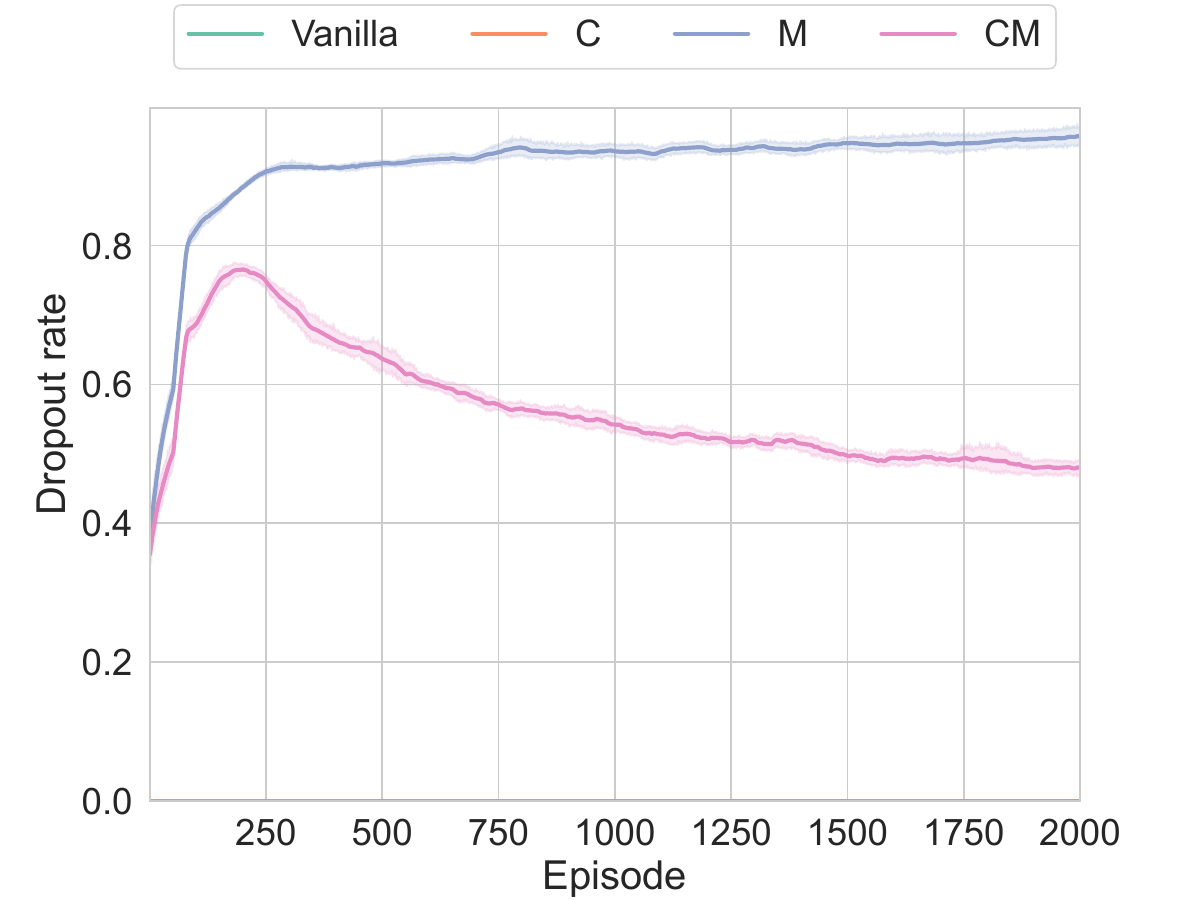}
        \subcaption{Reacher}
        \label{fig:erc_mining_Reacher}
    \end{subfigure}
    \begin{subfigure}[b]{0.32\linewidth}
        \centering
        \includegraphics[keepaspectratio=true,width=\linewidth]{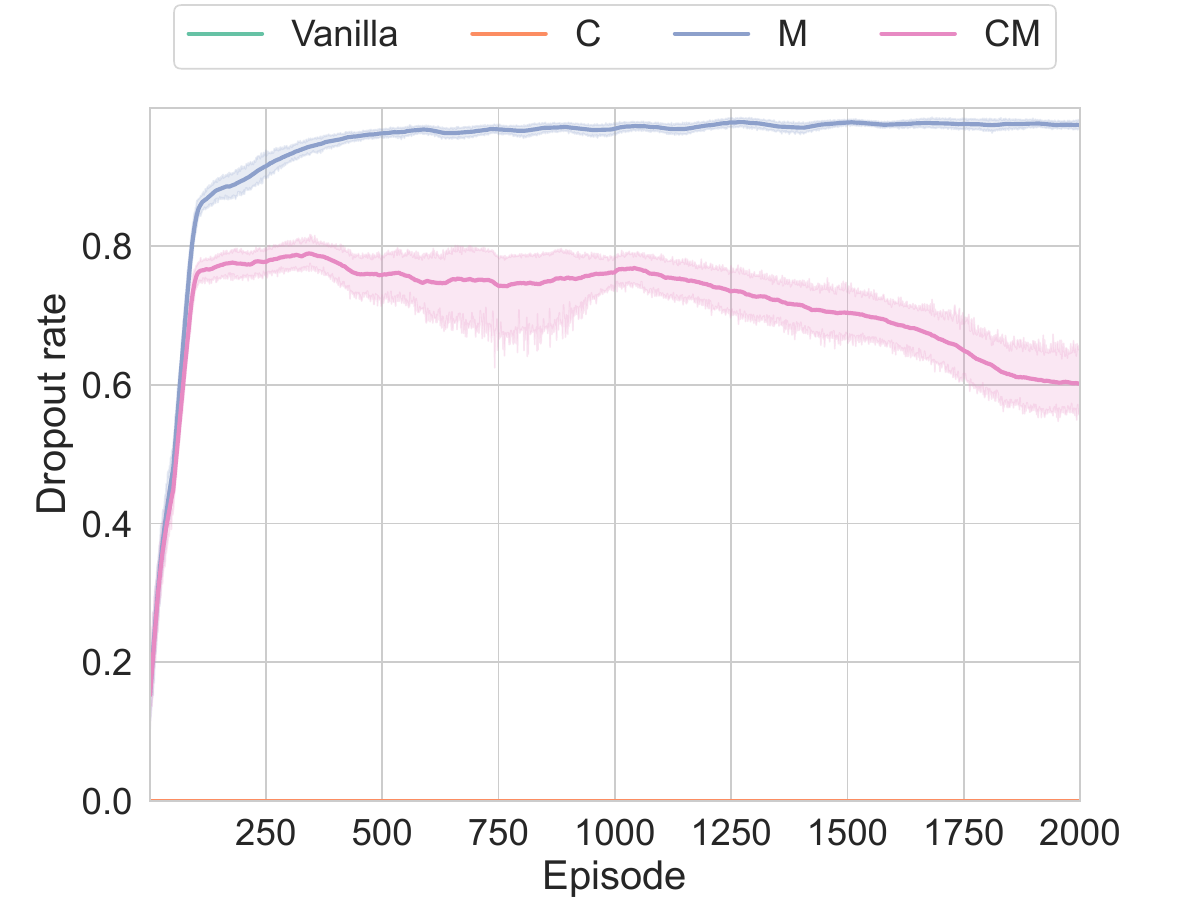}
        \subcaption{Hopper}
        \label{fig:erc_mining_Hopper}
    \end{subfigure}
    \begin{subfigure}[b]{0.32\linewidth}
        \centering
        \includegraphics[keepaspectratio=true,width=\linewidth]{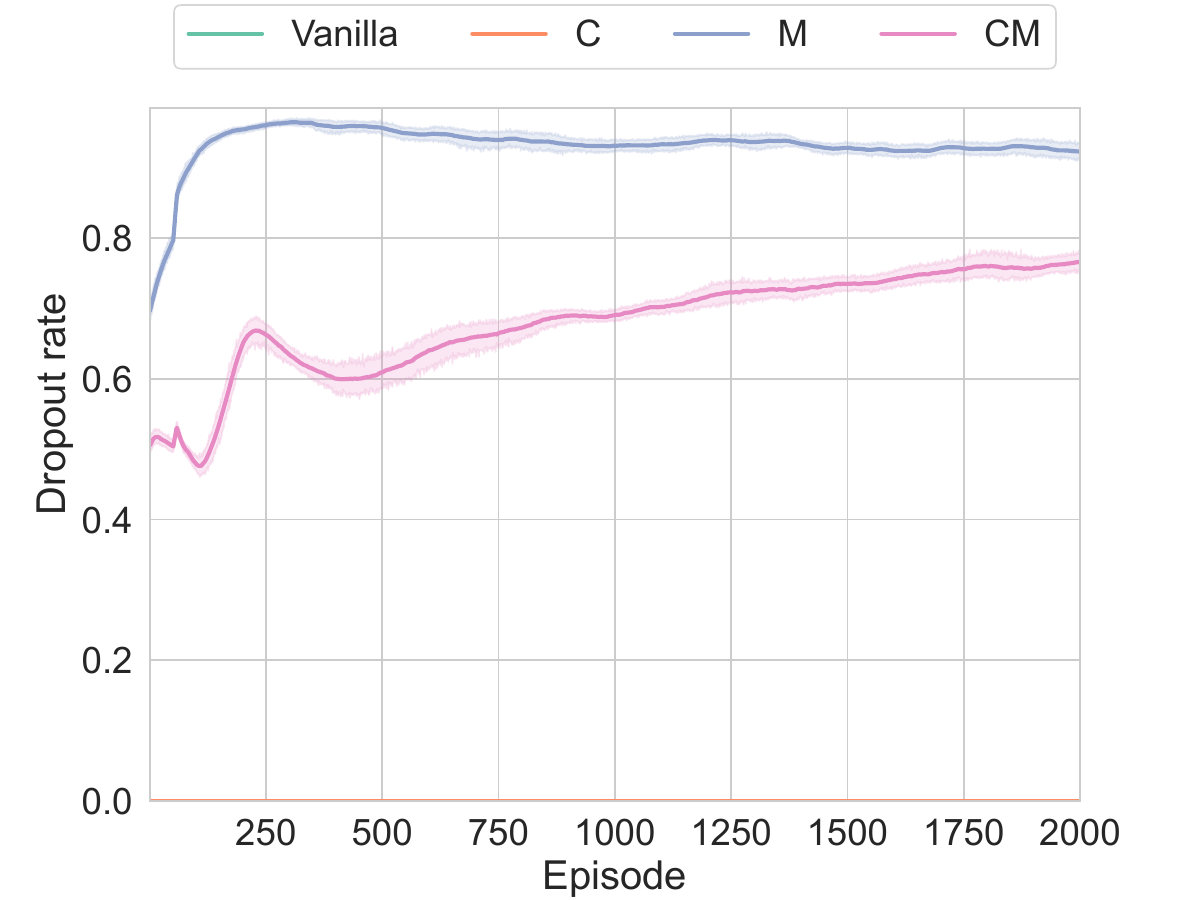}
        \subcaption{HalfCheetah}
        \label{fig:erc_mining_HalfCheetah}
    \end{subfigure}
    \caption{$p(M \mid s, a, b)$ in the mining trick:
    While the mining trick alone saturated the dropout rate and made it unavailable for replaying enough empirical data, the two tricks made the selection work properly without saturation.
    }
    \label{fig:erc_mining}
\end{figure}

Next, the needs for the two stabilization tricks are demonstrated through ablation tests.
Specifically, the presence or absence of the two stabilization tricks (labeled with their initials `C' and `M') is switched by setting $\eta_{C,M} = 0$.
The four conditions of that combination are compared in the following three tasks in Mujoco~\citep{todorov2012mujoco}: \textit{Reacher}; \textit{Hopper}; and \textit{HalfCheetah}.

The learning curves for these returns are shown in Fig.~\ref{fig:score}.
As can be seen from the results, Vanilla (a.k.a. the standard A2C) without the two stabilization tricks did not learn any tasks at all, probably because it does not satisfy ERC.
Adding the counteraction trick alone did not improve learning as well, but adding the mining trick improved somewhat.
Nevertheless, it seems that either of them does not satisfy ERC or learn the optimal policy.
On the other hand, only the case using the two stabilization tricks was able to learn all tasks.
From these results, it can be concluded that both of the two stabilization tricks are necessary (and sufficient) to satisfy ERC and learn the optimal policy.

Now let's see why both of the stabilization tricks are required.
To this end, the internal parameters for the respective stabilization tricks, $\omega(s, a, b)$ and $p(M \mid s, a, b)$, are depicted in Figs.~\ref{fig:erc_counteract} and~\ref{fig:erc_mining}.
One can find that the cases only with one of the stabilization tricks saturated the respective internal parameters immediately.
That is, if using the counteraction trick only, its regularization to $\pi \to b$ is not enough for satisfying ERC;
and if using the mining trick only, its selection is too strict for learning the optimal policy.
In contrast, by using both, the internal parameters converged to task-specific values without saturation.
That is why the two stabilization tricks need to be employed together to increase the amount of replayable empirical data to reach a level where the optimal policy can be learned while still satisfying ERC.

\subsection{Performance comparison}

\begin{figure}[tb]
    \begin{subfigure}[b]{0.32\linewidth}
        \centering
        \includegraphics[keepaspectratio=true,width=\linewidth]{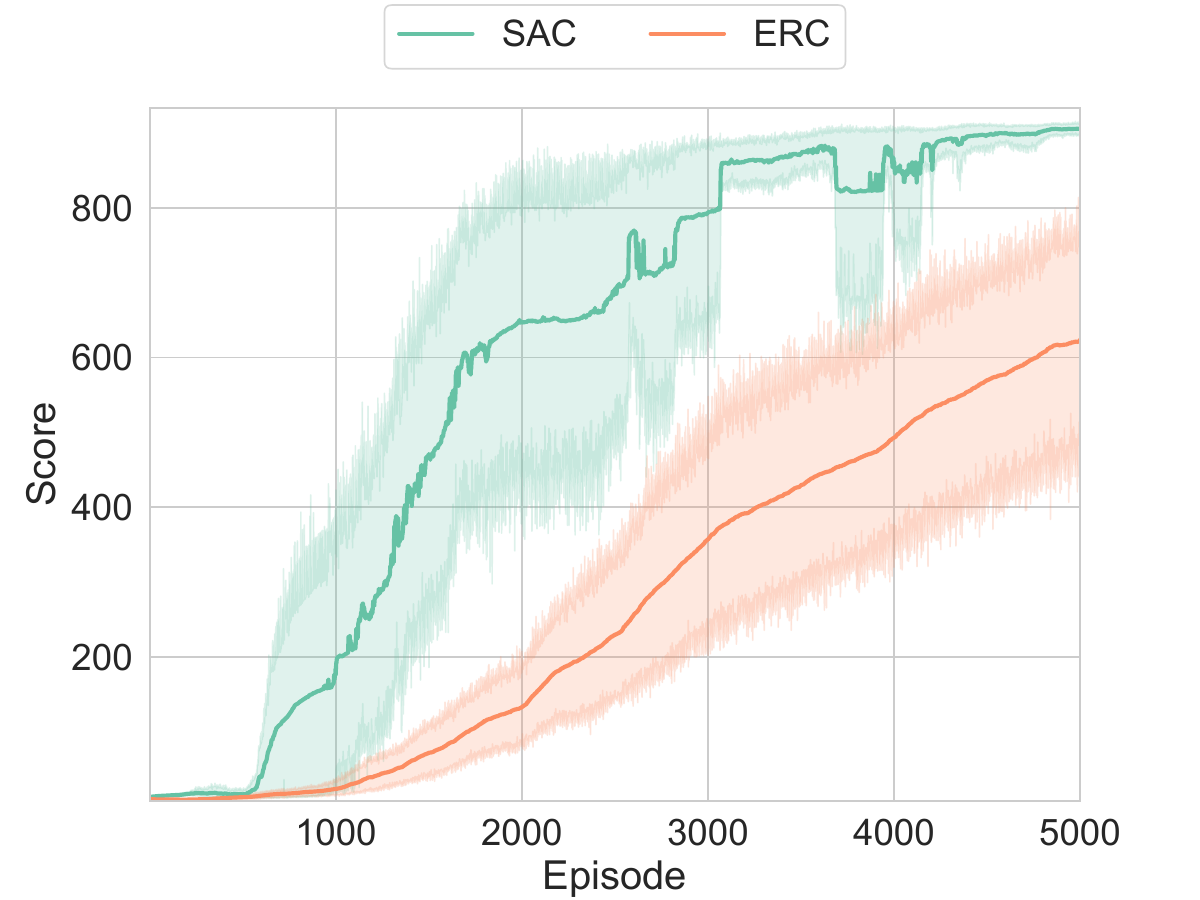}
        \subcaption{Quadruped}
        \label{fig:score_Quadruped}
    \end{subfigure}
    \begin{subfigure}[b]{0.32\linewidth}
        \centering
        \includegraphics[keepaspectratio=true,width=\linewidth]{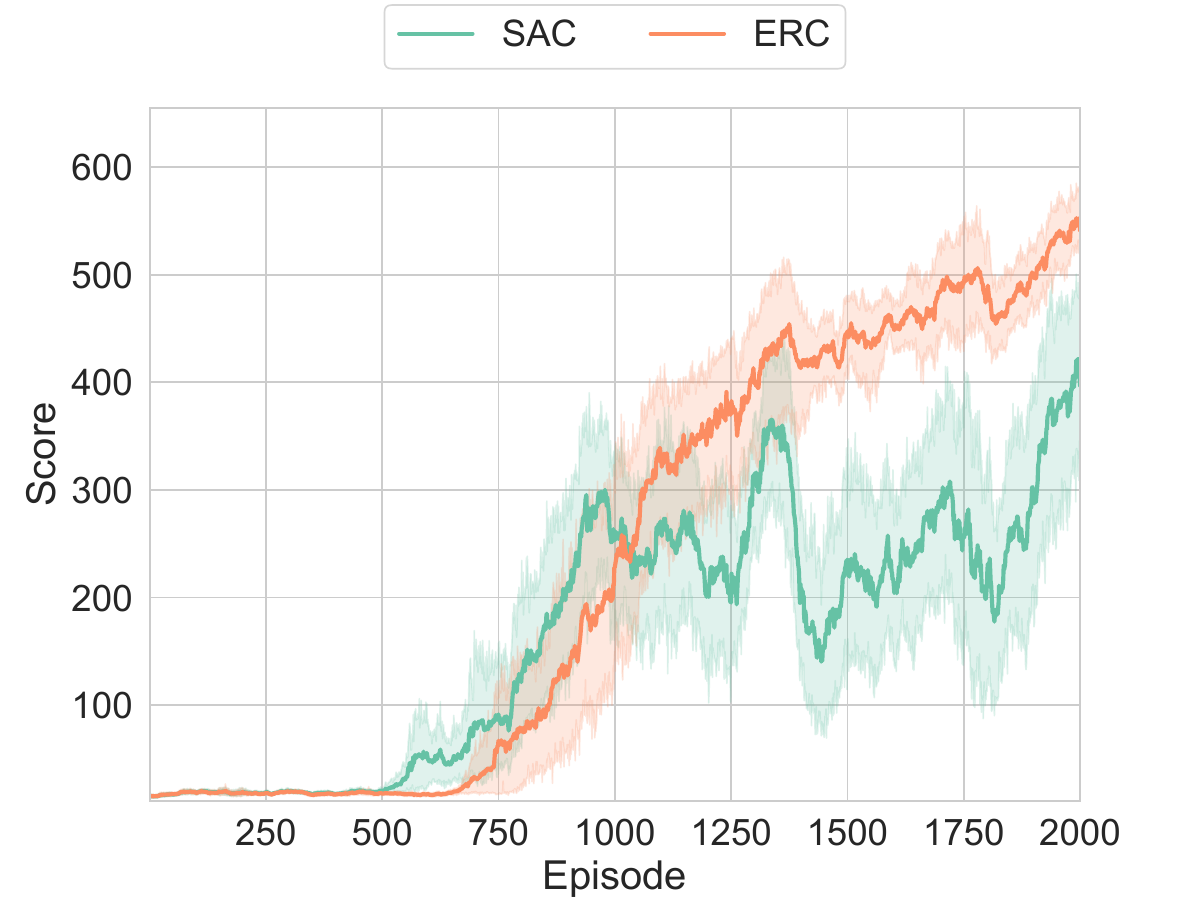}
        \subcaption{Swimmer}
        \label{fig:score_Swimmer}
    \end{subfigure}
    \begin{subfigure}[b]{0.32\linewidth}
        \centering
        \includegraphics[keepaspectratio=true,width=\linewidth]{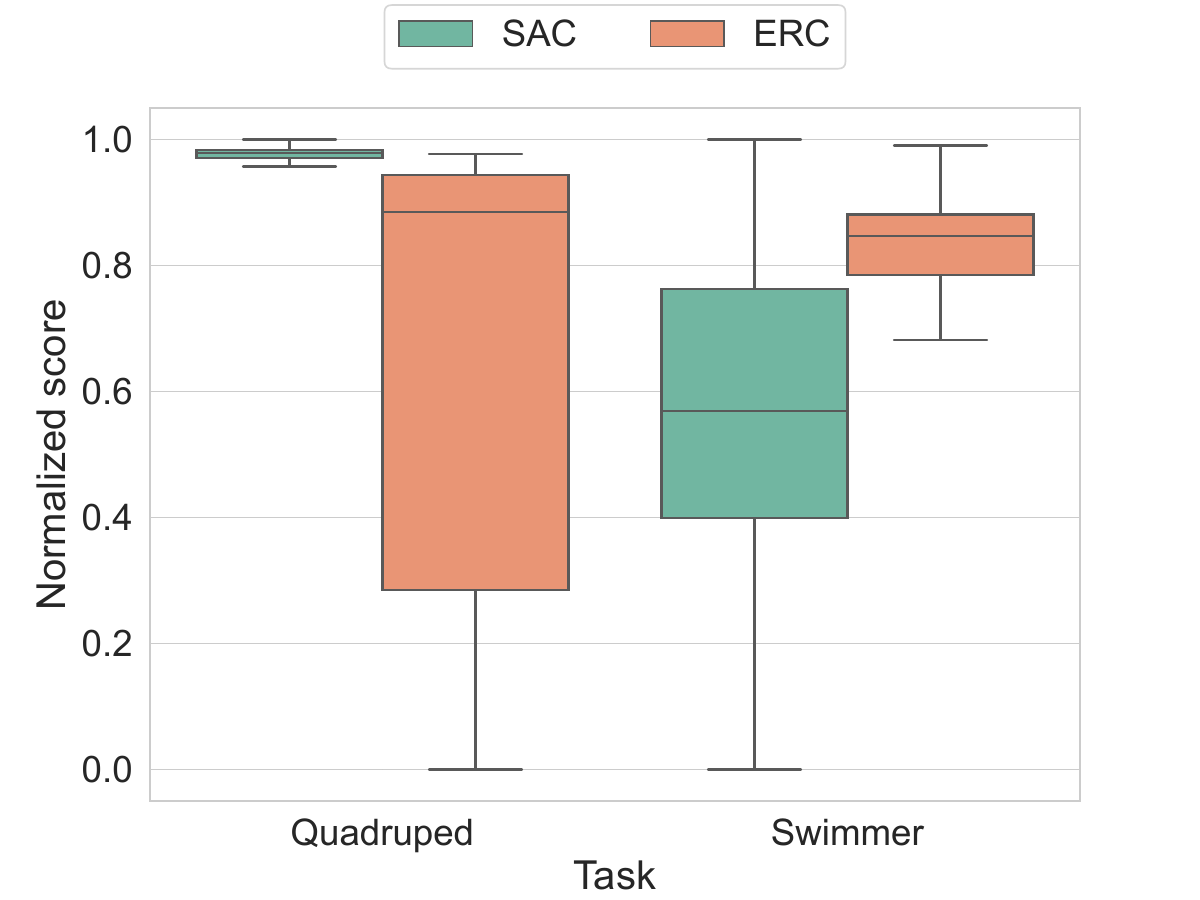}
        \subcaption{Returns after learning}
        \label{fig:test_practical}
    \end{subfigure}
    \caption{Returns on practical tasks:
    From (a) and (b), it can be found that SAC has faster learning speed, while ERC has more stable learning ability, resulting in (c) learning performance of ERC comparable to SAC.
    }
    \label{fig:practical}
\end{figure}

Finally, the learning performance of the A2C with the proposed stabilization tricks (labeled ERC) is compared to the state-of-the-art off-policy algorithm, SAC~\citep{haarnoja2018soft}.
Two practical tasks with more than 10-dimensional action space, \textit{QuadrupedWalk} (with $|\mathcal{A}| = 12$) and \textit{Swimmer15D} (with $|\mathcal{A}| = 14$) in dm\_control~\citep{tunyasuvunakool2020dm_control}, are solved.

The learning curves and test results after learning are depicted in Fig.~\ref{fig:practical}.
As can be seen, while SAC solved Quadruped at a high level, it did not solve Swimmer consistently.
On the other hand, ERC sometimes delayed convergence on Quadruped, but showed high performance on Swimmer.
In addition, the trends of the learning curves suggested that SAC learns faster, while ERC learns more stably.
Thus, it can be said that ERC has a learning performance comparable to that of SAC, although they have different strengths in different tasks.

As a remark, the difference in performance between ERC and SAC might be attributed to the difference in the policy gradients used.
Actually, ERC learns the policy with likelihood-ratio gradients, while SAC does so with reparameterization gradients~\citep{parmas2021unified}.
The former obtains the gradients over the entire (multi-dimensional) policy, which cannot optimize each component of actions independently.
As a result, this is useful to learn tasks that require synchonization of actions.
The latter, on the other hand, obtains the gradients for each component of actions, and thus can quickly learn tasks that do not require precise synchronization.
These learning characteristics correspond to the two types of tasks solved in this study.
Hence, it is suggested that SAC performed well for Quadruped, where synchronization at each leg is sufficient, and ERC performed well for Swimmer, where synchronization at all joints is needed.
Note that policy improvements by mixing them have been proposed, which would be expected to be useful~\citep{gu2017interpolated}.
The proposed stabilization tricks allow for the policy improvements combined with ER regardless of which gradient is used.
That would facilitate the development of such hybrid algorithms and improve learning performance in a complementary manner.

\section{Discussions}

\subsection{Relations to conventional algorithms}
\label{subsec:onpolicy}

As demonstrated in the above section, ERC hypothesized in this study can be achieved by limiting the replayed empirical data to ones acceptable by the applied RL algorithm.
At the same time, the improvement in learning efficiency due to ER can be increased by expanding the acceptable set of empirical data.
These were achieved by the two stabilization tricks proposed, with $\pi \simeq b$ being the key to both.
This key corresponds to the concept of \textit{on-policyness}, where the empirical data contained in the replay buffer can be regarded as generated from the current policy $\pi$~\citep{fedus2020revisiting}.
In other words, the stabilization tricks aimed to stabilize the triplet loss and consequently increased the on-policyness.
Conversely, the usefulness of increasing the on-policyness in the previous studies~\citep{novati2019remember,sinha2022experience} can be explained in terms of mitigating the instability factors hidden in the triplet loss.

Actually, ER can be applied to make A2C off-policy~\citep{degris2012off}.
To this end, the most important trick is to change the sampler by importance sampling, as described above.
This trick involves weighting the original loss function by $\pi/b$, which is unstable when $b \simeq 0$ (i.e. rare actions are sampled).
Several heuristics have been proposed to resolve this instability~\citep{wang2017sample,zhang2019generalized,fakoor2020p3o}.
Specifically, the most famous off-policy actor-critic algorithm with ER, ACER~\citep{wang2017sample}, modifies the update rules of not only the policy function but also the value function, while the proposed tricks indicated no need to modify the value function.
Although Geoff-PAC~\citep{zhang2019generalized} and P3O~\citep{fakoor2020p3o} only modify the policy improvements, the former is tedious and lacks the intuitive explanatory capability of ERC, while the latter is inappropriate for confirming ERC because it uses the freshly collected empirical data for learning separately from those sampled from ER.

Thus, previous studies have only focused on off-policyness and improved learning performance of the algorithms, without any discussion why ER is applicable.
On the other hand, this study explored ERC and implicitly obtained utility similar to making the target algorithm off-policy.
Indeed, the proposed stabilization tricks selectively replay the empirical data with $\pi \simeq b$ while increasing the number of acceptable ones.
As a result, the weight in this important sampling can be ignored as $\pi/b \simeq 1$.
That is why the A2C with the proposed stabilization tricks, although not explicitly off-policy, made ER available as if they were off-policy.

As a remark, the report that PPO (and its variants), which is regarded to be on-policy~\citep{schulman2017proximal}, could utilize ER empirically~\citep{kobayashi2023proximal} is related to the above considerations and the proposed stabilization tricks.
That is, PPO applies regularization aiming at $\pi \simeq b$, and if $\pi$ deviates from $b$, it clips the weight of the importance sampling, resulting in excluding that data.
Thus, PPO can increase the on-policyness of empirical data and exclude from replays that are unacceptable, so ER is actually applicable (see the empirical results in Appendix~\ref{app:ppo}).

Finally, another interpretation of why DDPG~\citep{lillicrap2016continuous}, SAC~\citep{haarnoja2018soft}, and TD3~\citep{fujimoto2018addressing}, which are considered off-policy, can stably learn the optimal policy using ER.
These algorithms optimize the current policy $\pi$ by using reparameterization gradients through the actions generated from $\pi$ fed into the action value function $Q$.
As mentioned in the SAC paper, this can be derived from the minimization of KL divergence between $\pi$ and the optimal policy based on $Q$ only.
In other words, unlike algorithms that use likelihood-ratio gradients such as A2C, they do not deal with the triplet loss.
Therefore, it can be considered that ER is applicable since there is no instability factors induced by the triplet loss in the first place.

\subsection{Limitations of stabiliation tricks}

For ERC holding, this paper proposed the two stabilization tricks, the contributions of which were evaluated in the above section.
However, these naturally leave room for improvements probably because they have been designed heuristically.
First, using them requires a discriminator $D$ learned through eq.~\eqref{eq:loss_discriminator}, which needs an extra computational cost.
While it is possible to implement the two stabilization tricks using only $d$ in eq.~\eqref{eq:density_ratio} without $D$, $d$ is an unstable variable, so a lightweight stabilizer will be needed to replace $D$.

It is also important to note that the decision on $D$ and $d$ is based on the likelihood $\pi$ and $b$.
That is, when the policy deals with a high-dimensional action space, even a small deviation in action may be judged as $\pi \neq b$ severely.
In fact, when the RL benchmark with musculoskeletal model, i.e. \textit{myosuite}~\citep{caggiano2022myosuite}, was tested with $|\mathcal{A}|=39$, $p(M \mid s, a, b)$ often converged to one depending on the conditions.
This means that the empirical data were rarely replayed.
To avoid this excessive exclusion of empirical data, it would be useful to either determine the final exclusion of empirical data by summarizing the respective judgements on individual action dimension separately; or to optimize the policy by masking the inappropriate action dimension only.

In addition, the counteraction alone did not satisfy ERC, as indicated in the ablation tests.
This is probably due to saturation of the regularization gain $\omega$ (see Fig.~\ref{fig:erc_counteract}).
Although a non-saturated gain made learning unstable empirically, it would be better to relax the saturation to some extent.
Alternatively, an auto-tuning trick, which is possible by once interpreting the regularization as the corresponding constraint~\citep{haarnoja2018soft2,kobayashi2023soft}, is considered to be useful.

\section{Conclusion}

This study reconsidered the factors that determine whether ER is applicable to RL algorithms.
To this end, the instability factors that ER might induce especially in on-policy algorithms were first revealed.
In other words, it was found that the policy gradient algorithms can be regarded as the minimization of triplet loss in metric learning, inheriting its instability factors.
To alleviate them, the two stabilization tricks, the counteraction and mining, were proposed as countermeasures attached to arbitrary RL algorithms.
The counteraction and mining are responsible for i) expanding the set of acceptable empirical data for each RL algorithm and ii) excluding empirical data outside the set, respectively.
Through multiple simulations, ERC indeed holded by using these two stabilization tricks.
Furthermore, the standard on-policy algorithm with them achieved the learning performance comparable to the state-of-the-art off-policy algorithm.

As described in the discussion, the two stabilization tricks proposed to satisfy ERC leave some room for improvements.
In particular, since the hypotheses formulated in this study is now deeply related to the on-policyness of the empirical data in the replay buffer, they would be improved based on this perspective.
Afterwards, other RL algorithms will be integrated with the stabilization tricks for further investigations of ERC.
On the other hand, although the simplest ER method was tested for simplicity since the proposed tricks allow for arbitrary ER methods, it would be interesting to consider more sophisticated methods~\citep{liu2022prioritized,wei2024re} and methods modified to suit the problem~\citep{christianos2020shared} or algorithm~\citep{banerjee2024improved}.
In particular, by following the latter direction and developing an ER method suitable for on-policy RL algorithms, a significant performance improvement might be expected.

\backmatter

%
%
%

\bmhead{Acknowledgments}

This research was supported by ``Strategic Research Projects'' grant from ROIS (Research Organization of Information and Systems).

\section*{Declarations}

\bmhead{Competing Interests}

The author declares that there is no known competing financial interests or personal relationships that could have appeared to influence the work reported in this paper.

\bmhead{Author Contributions}

Taisuke Kobayashi contributed to everything for this paper: Conceptualization, Methodology, Software, Validation, Investigation, Visualization, Funding acquisition, and Writing.

\bmhead{Compliance with Ethical Standards}

The data used in this study was exclusively generated by the author.
No research involving human participants or animals has been performed.

\bmhead{Data Availability}

The datasets generated during and/or analyzed during the current study are available from the corresponding author on reasonable request.

\begin{appendices}

\section{Details of implementation}
\label{app:impl}

The algorithms used in this paper is implemented with Pytorch~\citep{paszke2019pytorch}.
This implementation is based on the one in the literature~\citep{kobayashi2023intentionally}.
The characteristic hyperparameters in this implementation are listed up in Table~\ref{tab:param}.

The policy and value functions are approximated by the fully-connected neural networks consisting of two hidden layers with 100 neurons for each.
As activation functions, Squish function~\citep{barron2021squareplus,kobayashi2023design} and RMSNorm~\citep{zhang2019root} are combined.
AdaTerm~\citep{ilboudo2023adaterm}, which is the noise-robust optimizer, is adopted for robustly optimizing parameters against noises caused by bootstrapped learning in RL.
Similarly, the target networks that are updated by CAT-soft update~\citep{kobayashi2024consolidated} are employed to stabilize learning (and to prevent learning speed degradation).
In A2C, the target networks are applied to both the policy $\pi$ and the value function $V$, while in SAC they are applied only to the action value function $Q$.
In addition, A2C enhances output continuity by using L2C2~\citep{kobayashi2022l2c2} with default parameters, although SAC does not so due to reproduction of its standard implementation.

Both A2C and SAC policies are modeled as Student's t-distribution with high global exploration capability~\citep{kobayashi2019student}.
Therefore, the outputs from the networks are three model parameters: position, scale, and degrees of freedom.
However, as in the standard implementation of SAC, the process of converting the generated action to a bounded range is also explicitly considered as a probability distribution.
On the other hand, in A2C, this process is performed implicitly on the environment side and is not reflected in the probability distribution.

SAC approximates the two action value functions $Q_{1,2}$ with independent networks as in the standard implementation, and aims at conservative learning by selecting the smaller value.
On the other hand, A2C aims at stable learning by outputting 10 values from shared networks and using the median as the representative value.
In order to enhance the effect of ensemble learning, the outputs are computed with both learnable and unlearnable parameters, so that each output can easily take on different values (especially in unlearned regions)~\citep{osband2018randomized}.

A2C and SAC share the settings of ER, with a buffer size of 102,400 and a batch size of 256.
The replay buffer is in FIFO format that deletes the oldest empirical data when the buffer size is exceeded.
At the end of each episode, half of the empirical data stored in ER is replayed uniformly at random.

\begin{table*}[tb]
    \caption{Parameter configuration}
    \label{tab:param}
    \centering
    \begin{tabular}{ccc}
        \hline\hline
        Symbol & Meaning & Value
        \\
        \hline
        -- & \#Hidden layers & $2$
        \\
        -- & \#Neurons for each hidden layer & $100$
        \\
        -- & Activation function & Squish + RMSNorm
        \\
        \hline
        $\gamma$ & Discount factor & $0.99$
        \\
        $\alpha$ & Learning rate for AdaTerm & $10^{-3}$
        \\
        $\tau$ & Update rate of target networks & $0.1$
        \\
        \hline
        $|\mathcal{B}|$ & Buffer size & $102400$
        \\
        $N_b$ & Batch size & $256$
        \\
        $N_r$ & \#Replayed data & $|\mathcal{B}|/2$
        \\
        \hline\hline
    \end{tabular}
\end{table*}

\section{Application to PPO}
\label{app:ppo}

\begin{figure}[tb]
    \begin{subfigure}[b]{0.32\linewidth}
        \centering
        \includegraphics[keepaspectratio=true,width=\linewidth]{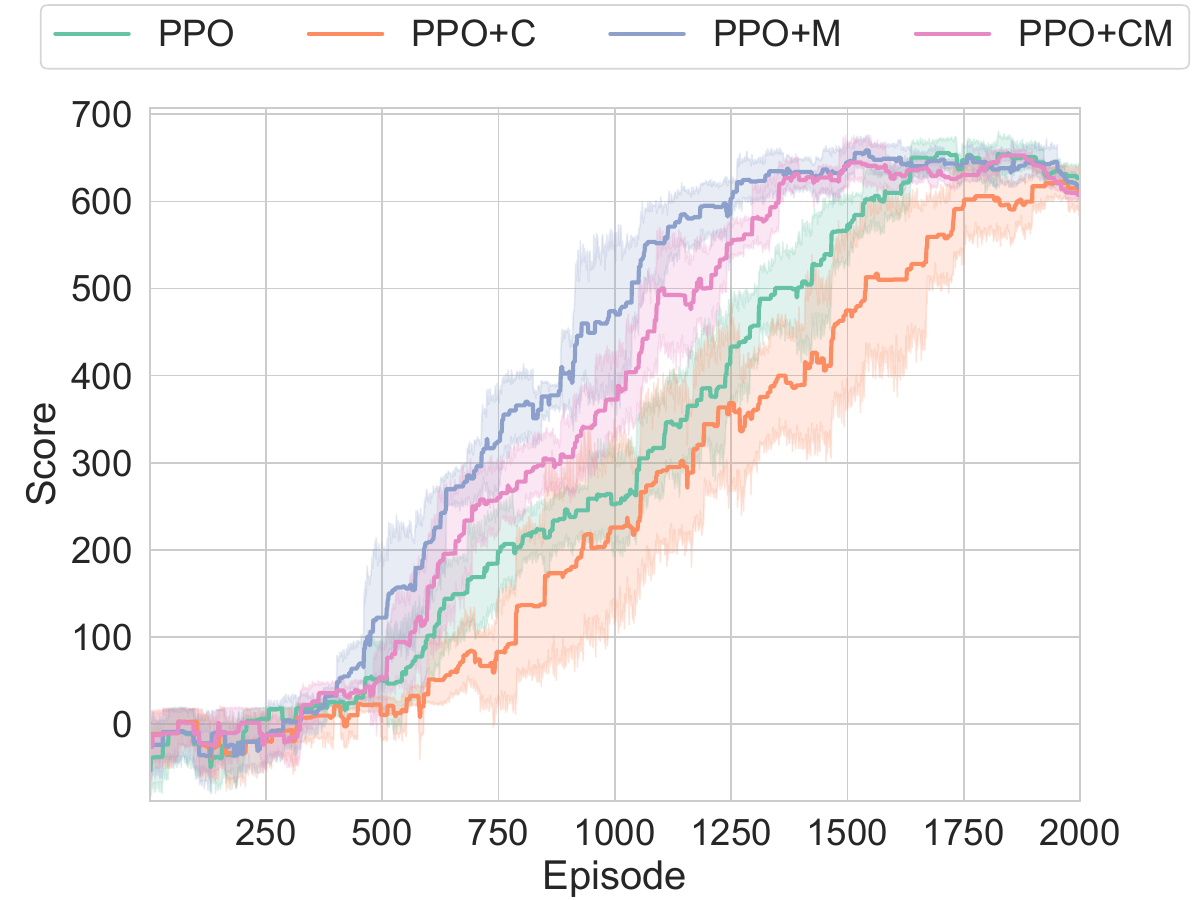}
        \subcaption{Quadrotor}
        \label{fig:score_Quadrotor}
    \end{subfigure}
    \begin{subfigure}[b]{0.32\linewidth}
        \centering
        \includegraphics[keepaspectratio=true,width=\linewidth]{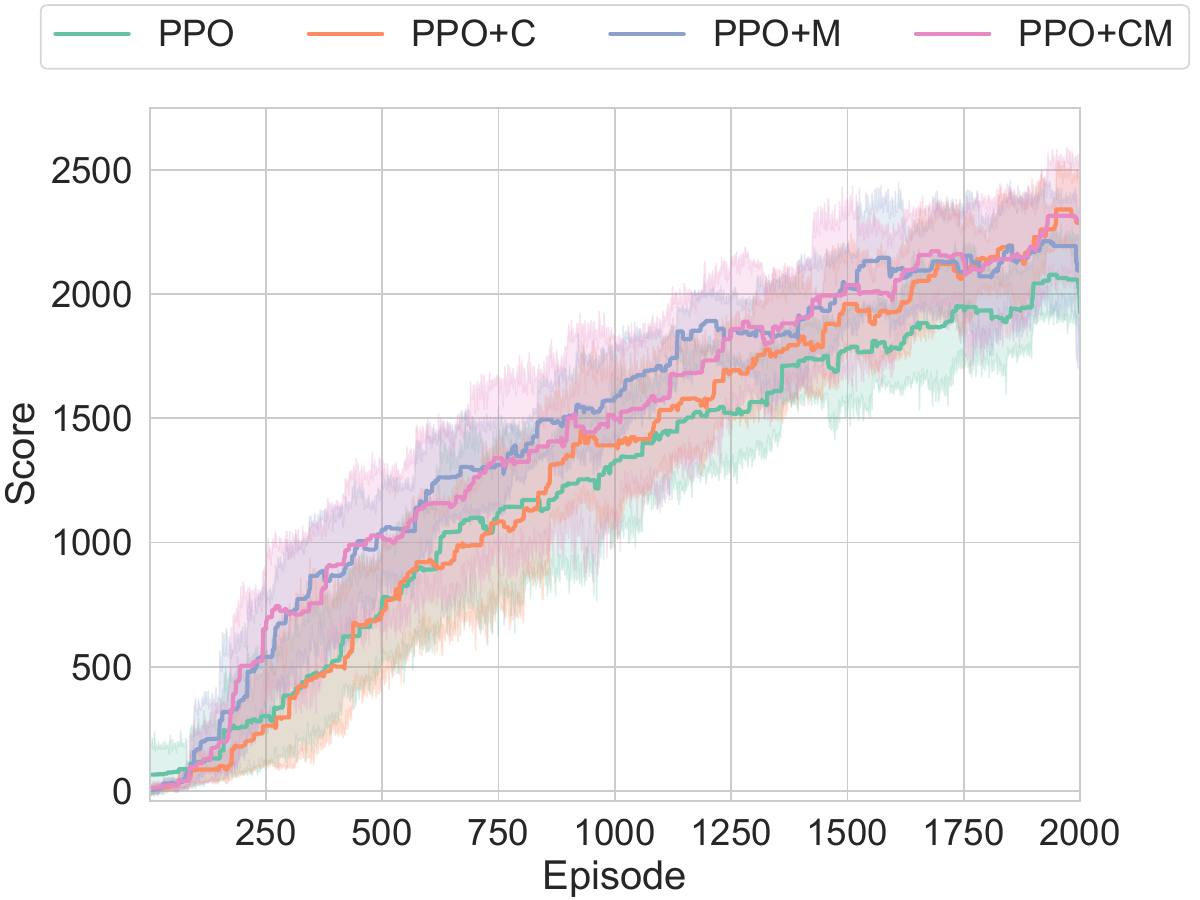}
        \subcaption{Fixedwing}
        \label{fig:score_Fixedwing}
    \end{subfigure}
    \begin{subfigure}[b]{0.32\linewidth}
        \centering
        \includegraphics[keepaspectratio=true,width=\linewidth]{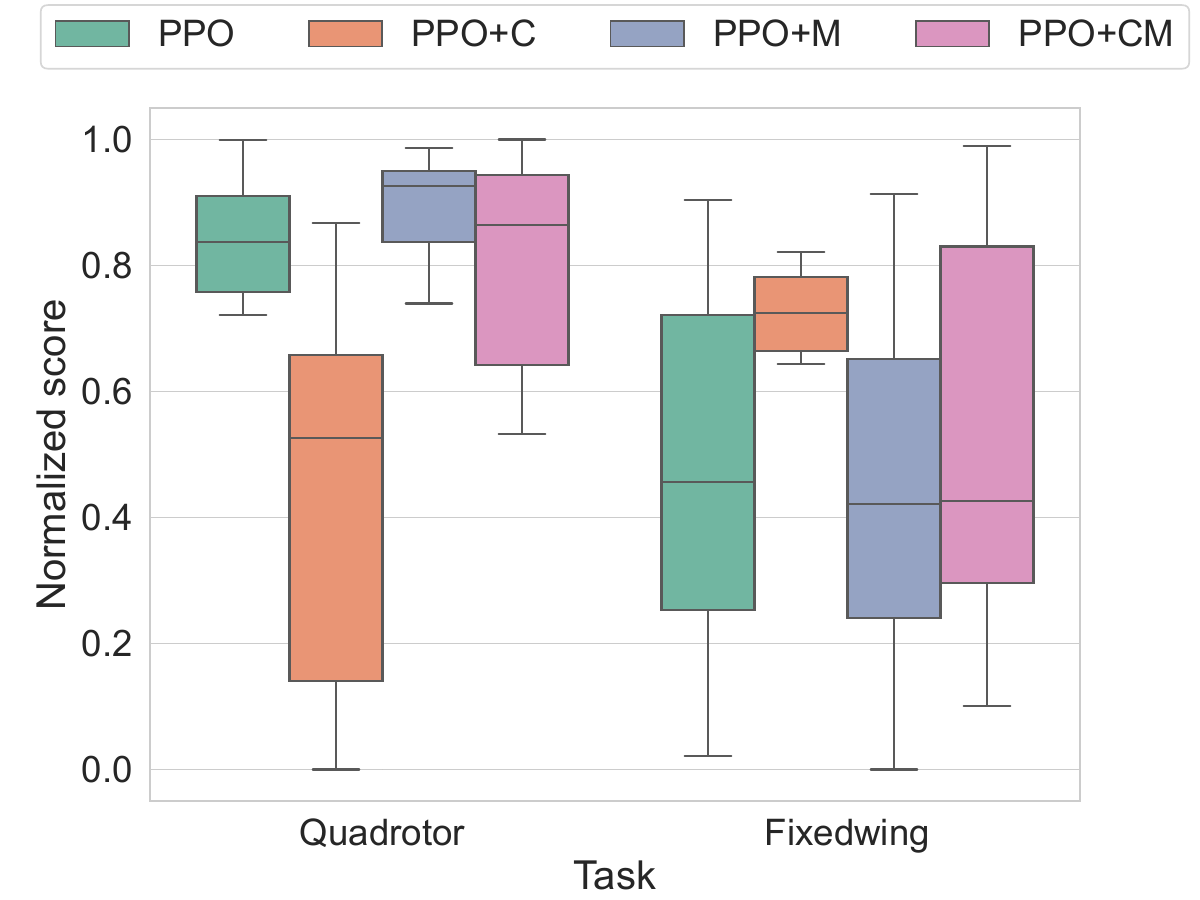}
        \subcaption{Returns after learning}
        \label{fig:test_ppo}
    \end{subfigure}
    \caption{Returns with PPO on PyFlyt:
    From (a) and (b), it can be found that the mining trick improves learning speed, while the counteraction trick facilitates exploration, which yielded the maximum and stable return on Fixedwing in (c).
    }
    \label{fig:ppo}
\end{figure}

The proposed stabilization tricks are applied to PPO~\citep{schulman2017proximal}, a latest on-policy algorithm other than A2C.
PPO multiplies the policy improvement by the policy likelihood ratio using importance sampling, and it is clipped to force the gradient to zero if it is excessive.
The recommended clipping threshold, $0.2$, is employed.
Since this paper uses an ER that accumulates empirical data for the simplest single transition, GAE~\citep{schulman2016high}, which is often used in conjunction with PPO, is ignored.
In addition, to check the regularization effects by the stabilization tricks, an extra regularization term, i.e. the policy entropy, is also omitted.
As mentioned in Introduction, as PPO is empirically ER-applicable, it is possible to quantitatively compare the learning trends and changes in the final policies due to each stabilization trick.

With the above setup, the results of solving \textit{QuadX-Waypoints-v2} and \textit{Fixedwing-Waypoints-v2} provided by PyFlyt~\citep{tai2023pyflyt} are summarized below.
Note that these tasks intend to control different drones (i.e. a quadrotor and a fixed-wing drone), respectively.
The learning curves for 7 trials and test results after learning in the four conditions diverging with and without the two stabilization tricks are depicted in Fig.~\ref{fig:ppo}.

First, the vanilla PPO shows the remarkable but expected results.
Although PPO is originally considered an on-policy algorithm, as described in the text, it achieved learning of both tasks in combination with ER even without the addition of the proposed stabilization tricks.
This is because, as discussed in Section~\ref{subsec:onpolicy}, PPO performs regularization and clipping heuristically such that $\pi \simeq b$ (i.e. on-policyness) holds.
In fact, PPO with only one of the stabilization trick did not saturate the corresponding internal parameter, indicating that ERC was satisfied without the lack of its functionality.
Note, however, that PPO by itself may not be sufficient to satisfy ERC, since GAE, which was omitted from the implementation this time, relies more heavily on the behavior policy than the simple advantage function (i.e. TD error).

Next, the contributions of the proposed stabilization tricks are confirmed.
First, it is easy to see that the mining trick increases learning speed, while the counteraction trick tends to decrease it.
This may be due to the fact that the direction of policy improvements becomes clearer by making the mining trick replay only the empirical data that are useful for learning, while the counteraction trick restricts policy improvements by binding $\pi \simeq b$.
For the former, actually, the scale of TD error, which implies the learning direction, was increased when the mining trick was added.

On the other hand, the counteraction trick seems to improve the exploration capability in exchange for the learning speed.
The return at the end of learning was maximized by including the counteraction trick on Fixedwing, although the return on Quadrotor with it was lower than others due to slow convergence.
This may be due to the increase in entropy of $\pi$ by regularizing it to various the behavior policies in the replay buffer.
In fact, the addition of the counteraction trick yielded the decrease of $\ln \pi$, which corresponds to the negative entropy, during learning.

\end{appendices}
\bibliography{biblio}

\begin{thebibliography}{65}
\providecommand{\natexlab}[1]{#1}
\providecommand{\url}[1]{{#1}}
\providecommand{\urlprefix}{URL }
\providecommand{\doi}[1]{\url{https://doi.org/#1}}
\providecommand{\eprint}[2][]{\url{#2}}
 \bibcommenthead

\bibitem[{Banerjee et~al(2024)Banerjee, Chen, and Noman}]{banerjee2024improved}
Banerjee C, Chen Z, Noman N (2024) Improved soft actor-critic: Mixing
  prioritized off-policy samples with on-policy experiences. IEEE Transactions
  on Neural Networks and Learning Systems 35(3):3121--3129

\bibitem[{Barron(2021)}]{barron2021squareplus}
Barron JT (2021) Squareplus: A softplus-like algebraic rectifier. arXiv
  preprint arXiv:211211687

\bibitem[{Bejjani et~al(2018{\natexlab{a}})Bejjani, Papallas, Leonetti, and
  Dogar}]{bejjani2018planning}
Bejjani W, Papallas R, Leonetti M, et~al (2018{\natexlab{a}}) Planning with a
  receding horizon for manipulation in clutter using a learned value function.
  \eprint{1803.08100}

\bibitem[{Bejjani et~al(2018{\natexlab{b}})Bejjani, Papallas, Leonetti, and
  Dogar}]{bejjani2018planning2}
Bejjani W, Papallas R, Leonetti M, et~al (2018{\natexlab{b}}) Planning with a
  receding horizon for manipulation in clutter using a learned value function.
  In: IEEE-RAS International Conference on Humanoid Robots, IEEE, pp 1--9

\bibitem[{Bellet et~al(2022)Bellet, Habrard, and Sebban}]{bellet2022metric}
Bellet A, Habrard A, Sebban M (2022) Metric learning. Springer Nature

\bibitem[{Caggiano et~al(2022)Caggiano, Wang, Durandau, Sartori, and
  Kumar}]{caggiano2022myosuite}
Caggiano V, Wang H, Durandau G, et~al (2022) Myosuite--a contact-rich
  simulation suite for musculoskeletal motor control. In: Learning for Dynamics
  and Control Conference, PMLR, pp 492--507

\bibitem[{Chen et~al(2021)Chen, Li, and Tomizuka}]{chen2021interpretable}
Chen J, Li SE, Tomizuka M (2021) Interpretable end-to-end urban autonomous
  driving with latent deep reinforcement learning. IEEE Transactions on
  Intelligent Transportation Systems 23(6):5068--5078

\bibitem[{Cheng et~al(2016)Cheng, Gong, Zhou, Wang, and
  Zheng}]{cheng2016person}
Cheng D, Gong Y, Zhou S, et~al (2016) Person re-identification by multi-channel
  parts-based cnn with improved triplet loss function. In: IEEE conference on
  computer vision and pattern recognition, pp 1335--1344

\bibitem[{Christianos et~al(2020)Christianos, Sch{\"a}fer, and
  Albrecht}]{christianos2020shared}
Christianos F, Sch{\"a}fer L, Albrecht S (2020) Shared experience actor-critic
  for multi-agent reinforcement learning. Advances in neural information
  processing systems 33:10707--10717

\bibitem[{Cui et~al(2021)Cui, Osaki, and Matsubara}]{cui2021autonomous}
Cui Y, Osaki S, Matsubara T (2021) Autonomous boat driving system using
  sample-efficient model predictive control-based reinforcement learning
  approach. Journal of Field Robotics 38(3):331--354

\bibitem[{Degris et~al(2012)Degris, White, and Sutton}]{degris2012off}
Degris T, White M, Sutton RS (2012) Off-policy actor-critic. In: International
  Conference on Machine Learning, pp 179--186

\bibitem[{Fakoor et~al(2020)Fakoor, Chaudhari, and Smola}]{fakoor2020p3o}
Fakoor R, Chaudhari P, Smola AJ (2020) {P3O}: Policy-on policy-off policy
  optimization. In: Uncertainty in Artificial Intelligence, PMLR, pp 1017--1027

\bibitem[{Fedus et~al(2020)Fedus, Ramachandran, Agarwal, Bengio, Larochelle,
  Rowland, and Dabney}]{fedus2020revisiting}
Fedus W, Ramachandran P, Agarwal R, et~al (2020) Revisiting fundamentals of
  experience replay. In: International Conference on Machine Learning, PMLR, pp
  3061--3071

\bibitem[{Fujimoto et~al(2018)Fujimoto, Hoof, and
  Meger}]{fujimoto2018addressing}
Fujimoto S, Hoof H, Meger D (2018) Addressing function approximation error in
  actor-critic methods. In: International conference on machine learning, PMLR,
  pp 1587--1596

\bibitem[{Ganin et~al(2016)Ganin, Ustinova, Ajakan, Germain, Larochelle,
  Laviolette, March, and Lempitsky}]{ganin2016domain}
Ganin Y, Ustinova E, Ajakan H, et~al (2016) Domain-adversarial training of
  neural networks. Journal of machine learning research 17(59):1--35

\bibitem[{Gu et~al(2017)Gu, Lillicrap, Turner, Ghahramani, Sch{\"o}lkopf, and
  Levine}]{gu2017interpolated}
Gu SS, Lillicrap T, Turner RE, et~al (2017) Interpolated policy gradient:
  Merging on-policy and off-policy gradient estimation for deep reinforcement
  learning. Advances in neural information processing systems 30:3849--3858

\bibitem[{Haarnoja et~al(2018{\natexlab{a}})Haarnoja, Zhou, Abbeel, and
  Levine}]{haarnoja2018soft}
Haarnoja T, Zhou A, Abbeel P, et~al (2018{\natexlab{a}}) Soft actor-critic:
  Off-policy maximum entropy deep reinforcement learning with a stochastic
  actor. In: International conference on machine learning, PMLR, pp 1861--1870

\bibitem[{Haarnoja et~al(2018{\natexlab{b}})Haarnoja, Zhou, Hartikainen,
  Tucker, Ha, Tan, Kumar, Zhu, Gupta, Abbeel, and Levine}]{haarnoja2018soft2}
Haarnoja T, Zhou A, Hartikainen K, et~al (2018{\natexlab{b}}) Soft actor-critic
  algorithms and applications. arXiv preprint arXiv:181205905

\bibitem[{Hambly et~al(2023)Hambly, Xu, and Yang}]{hambly2023recent}
Hambly B, Xu R, Yang H (2023) Recent advances in reinforcement learning in
  finance. Mathematical Finance 33(3):437--503

\bibitem[{Hansen et~al(2018)Hansen, Pritzel, Sprechmann, Barreto, and
  Blundell}]{hansen2018fast}
Hansen S, Pritzel A, Sprechmann P, et~al (2018) Fast deep reinforcement
  learning using online adjustments from the past. Advances in Neural
  Information Processing Systems 31:10590--10600

\bibitem[{Ilboudo et~al(2023)Ilboudo, Kobayashi, and
  Matsubara}]{ilboudo2023adaterm}
Ilboudo WEL, Kobayashi T, Matsubara T (2023) Adaterm: Adaptive t-distribution
  estimated robust moments for noise-robust stochastic gradient optimization.
  Neurocomputing 557:126692

\bibitem[{Kalashnikov et~al(2018)Kalashnikov, Irpan, Pastor, Ibarz, Herzog,
  Jang, Quillen, Holly, Kalakrishnan, Vanhoucke
  et~al}]{kalashnikov2018scalable}
Kalashnikov D, Irpan A, Pastor P, et~al (2018) Scalable deep reinforcement
  learning for vision-based robotic manipulation. In: Conference on Robot
  Learning, PMLR, pp 651--673

\bibitem[{Kapturowski et~al(2019)Kapturowski, Ostrovski, Quan, Munos, and
  Dabney}]{kapturowski2019recurrent}
Kapturowski S, Ostrovski G, Quan J, et~al (2019) Recurrent experience replay in
  distributed reinforcement learning. In: International Conference on Learning
  Representations

\bibitem[{Kobayashi(2019)}]{kobayashi2019student}
Kobayashi T (2019) Student-t policy in reinforcement learning to acquire global
  optimum of robot control. Applied Intelligence 49(12):4335--4347

\bibitem[{Kobayashi(2022{\natexlab{a}})}]{kobayashi2022l2c2}
Kobayashi T (2022{\natexlab{a}}) L2c2: Locally lipschitz continuous constraint
  towards stable and smooth reinforcement learning. In: IEEE/RSJ International
  Conference on Intelligent Robots and Systems, IEEE, pp 4032--4039

\bibitem[{Kobayashi(2022{\natexlab{b}})}]{kobayashi2022optimistic}
Kobayashi T (2022{\natexlab{b}}) Optimistic reinforcement learning by forward
  kullback--leibler divergence optimization. Neural Networks 152:169--180

\bibitem[{Kobayashi(2023{\natexlab{a}})}]{kobayashi2023intentionally}
Kobayashi T (2023{\natexlab{a}}) Intentionally-underestimated value function at
  terminal state for temporal-difference learning with mis-designed reward.
  arXiv preprint arXiv:230812772

\bibitem[{Kobayashi(2023{\natexlab{b}})}]{kobayashi2023proximal}
Kobayashi T (2023{\natexlab{b}}) Proximal policy optimization with adaptive
  threshold for symmetric relative density ratio. Results in Control and
  Optimization 10:100192

\bibitem[{Kobayashi(2023{\natexlab{c}})}]{kobayashi2023soft}
Kobayashi T (2023{\natexlab{c}}) Soft actor-critic algorithm with
  truly-satisfied inequality constraint. arXiv preprint arXiv:230304356

\bibitem[{Kobayashi(2024)}]{kobayashi2024consolidated}
Kobayashi T (2024) Consolidated adaptive t-soft update for deep reinforcement
  learning. In: IEEE World Congress on Computational Intelligence, (to appear)

\bibitem[{Kobayashi and Aotani(2023)}]{kobayashi2023design}
Kobayashi T, Aotani T (2023) Design of restricted normalizing flow towards
  arbitrary stochastic policy with computational efficiency. Advanced Robotics
  37(12):719--736

\bibitem[{Levine(2018)}]{levine2018reinforcement}
Levine S (2018) Reinforcement learning and control as probabilistic inference:
  Tutorial and review. arXiv preprint arXiv:180500909

\bibitem[{Lillicrap et~al(2016)Lillicrap, Hunt, Pritzel, Heess, Erez, Tassa,
  Silver, and Wierstra}]{lillicrap2016continuous}
Lillicrap TP, Hunt JJ, Pritzel A, et~al (2016) Continuous control with deep
  reinforcement learning. In: International Conference on Learning
  Representations

\bibitem[{Lin(1992)}]{lin1992self}
Lin LJ (1992) Self-improving reactive agents based on reinforcement learning,
  planning and teaching. Machine learning 8(3-4):293--321

\bibitem[{Liu et~al(2022)Liu, Zhu, Jiang, Ye, and Zhao}]{liu2022prioritized}
Liu X, Zhu T, Jiang C, et~al (2022) Prioritized experience replay based on
  multi-armed bandit. Expert Systems with Applications 189:116023

\bibitem[{Mnih et~al(2015)Mnih, Kavukcuoglu, Silver, Rusu, Veness, Bellemare,
  Graves, Riedmiller, Fidjeland, Ostrovski et~al}]{mnih2015human}
Mnih V, Kavukcuoglu K, Silver D, et~al (2015) Human-level control through deep
  reinforcement learning. nature 518(7540):529--533

\bibitem[{Mnih et~al(2016)Mnih, Badia, Mirza, Graves, Lillicrap, Harley,
  Silver, and Kavukcuoglu}]{mnih2016asynchronous}
Mnih V, Badia AP, Mirza M, et~al (2016) Asynchronous methods for deep
  reinforcement learning. In: International conference on machine learning,
  PMLR, pp 1928--1937

\bibitem[{Novati and Koumoutsakos(2019)}]{novati2019remember}
Novati G, Koumoutsakos P (2019) Remember and forget for experience replay. In:
  International Conference on Machine Learning, PMLR, pp 4851--4860

\bibitem[{Oh et~al(2021)Oh, Rho, Moon, Son, Lee, and Chung}]{oh2021creating}
Oh I, Rho S, Moon S, et~al (2021) Creating pro-level ai for a real-time
  fighting game using deep reinforcement learning. IEEE Transactions on Games
  14(2):212--220

\bibitem[{Osband et~al(2018)Osband, Aslanides, and
  Cassirer}]{osband2018randomized}
Osband I, Aslanides J, Cassirer A (2018) Randomized prior functions for deep
  reinforcement learning. Advances in Neural Information Processing Systems
  31:8626--8638

\bibitem[{Parmas and Sugiyama(2021)}]{parmas2021unified}
Parmas P, Sugiyama M (2021) A unified view of likelihood ratio and
  reparameterization gradients. In: International Conference on Artificial
  Intelligence and Statistics, PMLR, pp 4078--4086

\bibitem[{Paszke et~al(2019)Paszke, Gross, Massa, Lerer, Bradbury, Chanan,
  Killeen, Lin, Gimelshein, Antiga et~al}]{paszke2019pytorch}
Paszke A, Gross S, Massa F, et~al (2019) Pytorch: An imperative style,
  high-performance deep learning library. Advances in neural information
  processing systems 32:8026--8037

\bibitem[{Saglam et~al(2023)Saglam, Mutlu, Cicek, and Kozat}]{saglam2023actor}
Saglam B, Mutlu FB, Cicek DC, et~al (2023) Actor prioritized experience replay.
  Journal of Artificial Intelligence Research 78:639--672

\bibitem[{Schaul et~al(2016)Schaul, Quan, Antonoglou, and
  Silver}]{schaul2016prioritized}
Schaul T, Quan J, Antonoglou I, et~al (2016) Prioritized experience replay. In:
  International Conference on Learning Representations

\bibitem[{Schroff et~al(2015)Schroff, Kalenichenko, and
  Philbin}]{schroff2015facenet}
Schroff F, Kalenichenko D, Philbin J (2015) Facenet: A unified embedding for
  face recognition and clustering. In: IEEE conference on computer vision and
  pattern recognition, pp 815--823

\bibitem[{Schulman et~al(2016)Schulman, Moritz, Levine, Jordan, and
  Abbeel}]{schulman2016high}
Schulman J, Moritz P, Levine S, et~al (2016) High-dimensional continuous
  control using generalized advantage estimation. In: International Conference
  on Learning Representations

\bibitem[{Schulman et~al(2017)Schulman, Wolski, Dhariwal, Radford, and
  Klimov}]{schulman2017proximal}
Schulman J, Wolski F, Dhariwal P, et~al (2017) Proximal policy optimization
  algorithms. arXiv preprint arXiv:170706347

\bibitem[{Sinha et~al(2022)Sinha, Song, Garg, and Ermon}]{sinha2022experience}
Sinha S, Song J, Garg A, et~al (2022) Experience replay with likelihood-free
  importance weights. In: Learning for Dynamics and Control Conference, PMLR,
  pp 110--123

\bibitem[{Srivastava et~al(2014)Srivastava, Hinton, Krizhevsky, Sutskever, and
  Salakhutdinov}]{srivastava2014dropout}
Srivastava N, Hinton G, Krizhevsky A, et~al (2014) Dropout: a simple way to
  prevent neural networks from overfitting. The journal of machine learning
  research 15(1):1929--1958

\bibitem[{Stooke et~al(2020)Stooke, Achiam, and Abbeel}]{stooke2020responsive}
Stooke A, Achiam J, Abbeel P (2020) Responsive safety in reinforcement learning
  by pid lagrangian methods. In: International Conference on Machine Learning,
  PMLR, pp 9133--9143

\bibitem[{Sutton and Barto(2018)}]{sutton2018reinforcement}
Sutton RS, Barto AG (2018) Reinforcement learning: An introduction. MIT press

\bibitem[{Tai et~al(2023)Tai, Wong, Innocente, Horri, Brusey, and
  Phang}]{tai2023pyflyt}
Tai JJ, Wong J, Innocente M, et~al (2023) Pyflyt--uav simulation environments
  for reinforcement learning research. arXiv preprint arXiv:230401305

\bibitem[{Todorov et~al(2012)Todorov, Erez, and Tassa}]{todorov2012mujoco}
Todorov E, Erez T, Tassa Y (2012) Mujoco: A physics engine for model-based
  control. In: IEEE/RSJ international conference on intelligent robots and
  systems, IEEE, pp 5026--5033

\bibitem[{Tunyasuvunakool et~al(2020)Tunyasuvunakool, Muldal, Doron, Liu,
  Bohez, Merel, Erez, Lillicrap, Heess, and
  Tassa}]{tunyasuvunakool2020dm_control}
Tunyasuvunakool S, Muldal A, Doron Y, et~al (2020) dm\_control: Software and
  tasks for continuous control. Software Impacts 6:100022

\bibitem[{Van~Seijen et~al(2009)Van~Seijen, Van~Hasselt, Whiteson, and
  Wiering}]{van2009theoretical}
Van~Seijen H, Van~Hasselt H, Whiteson S, et~al (2009) A theoretical and
  empirical analysis of expected sarsa. In: IEEE symposium on adaptive dynamic
  programming and reinforcement learning, IEEE, pp 177--184

\bibitem[{Wang et~al(2014)Wang, Song, Leung, Rosenberg, Wang, Philbin, Chen,
  and Wu}]{wang2014learning}
Wang J, Song Y, Leung T, et~al (2014) Learning fine-grained image similarity
  with deep ranking. In: IEEE conference on computer vision and pattern
  recognition, pp 1386--1393

\bibitem[{Wang et~al(2021)Wang, Song, Qi, Peng, Tang, Zhang, Li, Pi, He, Gao
  et~al}]{wang2021scc}
Wang X, Song J, Qi P, et~al (2021) Scc: An efficient deep reinforcement
  learning agent mastering the game of starcraft ii. In: International
  conference on machine learning, PMLR, pp 10905--10915

\bibitem[{Wang et~al(2017)Wang, Bapst, Heess, Mnih, Munos, Kavukcuoglu, and
  de~Freitas}]{wang2017sample}
Wang Z, Bapst V, Heess N, et~al (2017) Sample efficient actor-critic with
  experience replay. In: International Conference on Learning Representations

\bibitem[{Wei et~al(2024)Wei, Wang, Li, and Liang}]{wei2024re}
Wei W, Wang D, Li L, et~al (2024) Re-attentive experience replay in off-policy
  reinforcement learning. Machine Learning pp 1--23

\bibitem[{Wu et~al(2023)Wu, Escontrela, Hafner, Abbeel, and
  Goldberg}]{wu2023daydreamer}
Wu P, Escontrela A, Hafner D, et~al (2023) Daydreamer: World models for
  physical robot learning. In: Conference on Robot Learning, PMLR, pp
  2226--2240

\bibitem[{Xuan et~al(2020)Xuan, Stylianou, Liu, and Pless}]{xuan2020hard}
Xuan H, Stylianou A, Liu X, et~al (2020) Hard negative examples are hard, but
  useful. In: European Conference on Computer Vision, pp 126--142

\bibitem[{Yu et~al(2018)Yu, Liu, Gong, Ding, and Tao}]{yu2018correcting}
Yu B, Liu T, Gong M, et~al (2018) Correcting the triplet selection bias for
  triplet loss. In: European Conference on Computer Vision, pp 71--87

\bibitem[{Zhang and Sennrich(2019)}]{zhang2019root}
Zhang B, Sennrich R (2019) Root mean square layer normalization. Advances in
  Neural Information Processing Systems 32:12381--12392

\bibitem[{Zhang et~al(2019)Zhang, Boehmer, and Whiteson}]{zhang2019generalized}
Zhang S, Boehmer W, Whiteson S (2019) Generalized off-policy actor-critic.
  Advances in neural information processing systems 32:2001--2011

\bibitem[{Zhao et~al(2016)Zhao, Wang, Shao, and Zhu}]{zhao2016deep}
Zhao D, Wang H, Shao K, et~al (2016) Deep reinforcement learning with
  experience replay based on sarsa. In: IEEE symposium series on computational
  intelligence, IEEE, pp 1--6

\end{thebibliography}

\end{document}